\documentclass[a4paper,fleqn, review]{cas-dc}

\usepackage[square,sort&compress,comma,numbers]{natbib}

\usepackage{enumitem}
\usepackage{booktabs}
\usepackage{multirow}
\usepackage{threeparttable}
\usepackage{tabularray}

\usepackage{algorithm}
\usepackage{algpseudocode}

\usepackage{xcolor}

\usepackage{tikz}

\usepackage{cleveref}

\UseTblrLibrary{booktabs}
\UseTblrLibrary{functional}
\IgnoreSpacesOn
    \prgNewFunction \funcColor { m m m m m m m m } {
        \intStepOneInline {#7} {\arabic{rowcount}} {
            \intStepOneInline {#8} {\arabic{colcount}} {
                \fpSet \lTmpaFp {\cellGetText {##1} {####1}}
                \fpCompareTF {\lTmpaFp} < {#3} {
                    \tlSet \lTmpaTl {
                        \fpEval {
                            round(100 * (\lTmpaFp - #1) / (#3 - #1), 2)
                        }
                    }
                    \cellSetStyle {##1} {####1} {bg=#4!\lTmpaTl!#2}
                }{
                    \tlSet \lTmpaTl {
                        \fpEval {
                            round(100 * (\lTmpaFp - #3) / (#5 - #3), 2)
                        }
                    }
                    \cellSetStyle {##1} {####1} {bg=#6!\lTmpaTl!#4}
                }
            }
        }
    }
\IgnoreSpacesOff

\usepackage{caption} 
\newcolumntype{H}{>{\setbox0=\hbox\bgroup}c<{\egroup}@{}}

\def\tsc#1{\csdef{#1}{\textsc{\lowercase{#1}}\xspace}}
\tsc{WGM}
\tsc{QE}

\begin{document}
\let\WriteBookmarks\relax
\def\floatpagepagefraction{1}
\def\textpagefraction{.001}

\shorttitle{Evaluating ML-Based Anomaly Detection: A Study}    

\shortauthors{A.~Pekar and R.~Jozsa}  

\title [mode = title]{Evaluating ML-Based Anomaly Detection Across Datasets of Varied Integrity: A Case Study}  



%

\author[1,2]{Adrian Pekar}[type=author,
                         orcid=0000-0003-4511-8267]
\cormark[1]
\ead{apekar@hit.bme.hu}
\credit{
conceptualization,
methodology,
software,
validation,
formal analysis,
investigation,
resources,
data curation,
writing - original draft,
writing - review \& editing,
visualization,
supervision,
project administration}

\author[1]{Richard Jozsa}[type=author,
                         ]
\ead{jozsa.richard@edu.bme.hu}
\credit{
methodology,
formal analysis,
investigation,
data curation,
visualization}



\affiliation[1]{organization={
            Budapest University of Technology and Economics},
            addressline={M\H{u}egyetem rkp. 3.}, 
            city={Budapest},
            citysep={}, 
            postcode={H-1111}, 
            country={Hungary}}

\affiliation[2]{organization={
            HUN-REN-BME Information Systems Research Group},
            addressline={Magyar Tud\'{o}sok krt. 2}, 
            city={Budapest},
            citysep={}, 
            postcode={1117}, 
            country={Hungary}}

\cortext[1]{Corresponding author}



\begin{abstract}
Cybersecurity remains a critical challenge in the digital age, with network traffic flow anomaly detection being a key pivotal instrument in the fight against cyber threats. In this study, we address the prevalent issue of data integrity in network traffic datasets, which are instrumental in developing machine learning (ML) models for anomaly detection. We introduce two refined versions of the CICIDS-2017 dataset, NFS-2023-nTE and NFS-2023-TE, processed using NFStream to ensure methodologically sound flow expiration and labeling. Our research contrasts the performance of the Random Forest (RF) algorithm across the original CICIDS-2017, its refined counterparts WTMC-2021 and CRiSIS-2022, and our NFStream-generated datasets, in both binary and multi-class classification contexts.
We observe that the RF model exhibits exceptional robustness, achieving consistent high-performance metrics irrespective of the underlying dataset quality, which prompts a critical discussion on the actual impact of data integrity on ML efficacy. 
Our study underscores the importance of continual refinement and methodological rigor in dataset generation for network security research. As the landscape of network threats evolves, so must the tools and techniques used to detect and analyze them.
\end{abstract}




\maketitle

\begin{tikzpicture}[remember picture,overlay]
\node[anchor=north, align=center, text=red, font=\small, yshift=-.6cm] at (current page.north) {This version of the paper was accepted for publication in a future issue of the Computer Networks journal on 23 June 2024.};
\end{tikzpicture}

\section{Introduction}
\label{sec:intro}

Cybersecurity stands at the forefront of technological challenges in the digital age. As security threats evolve in complexity and volume, the need for advanced anomaly detection systems becomes imperative. Machine Learning (ML) and Artificial Intelligence (AI) have emerged as powerful allies in this fight, equipping us with sophisticated tools to anticipate, identify, and neutralize threats. At the core of these tools lies a dependency on high-quality datasets that accurately reflect the complexities of network traffic and cyber attacks.

Over the past decades, numerous datasets have been published to advance research in network traffic flow anomaly detection and cybersecurity. Prominent examples include datasets from the Canadian Institute for Cybersecurity, which offer extensive labeled traffic patterns~\cite{CIC}. The UNSW-NB15 dataset~\cite{UNSW-NB15} is also widely utilized, providing rich features extracted from real-world network traffic. Recently, new datasets such as LITNET-2020~\cite{electronics9050800} have emerged, offering annotated real-world network flow data capturing a wide array of attack types and normal traffic. Additionally, HIKARI-2021~\cite{app11177868} focuses on encrypted synthetic attacks and benign traffic to reflect modern network conditions. Several works also compile and list the available datasets in the field, including those mentioned above~\cite{10.1007/978-3-030-72802-1_9,10.1145/3441140,9165817}. Collectively, these datasets enhance the development and benchmarking of machine learning models in detecting network anomalies, ensuring a robust defense against evolving cyber threats.

The CICIDS-2017 dataset~\cite{CICIDS2017} has been pivotal in the development of methodologies for detecting network traffic anomalies. It offers a broad array of labeled traffic patterns and serve as a fundamental testing ground for a variety of ML-based approaches. Despite its widespread adoption and significant contributions to the field, recent scrutiny has exposed errors within these datasets that could skew the performance of detection systems trained on them~\cite{engelen2021,engelen2022,lanvin2023}. These inaccuracies highlight a critical need for continuous validation and refinement of measurement techniques to ensure that the models are learning from data that accurately represents real-world conditions.

This paper delves into a thorough examination and comparative analysis of five datasets---CICIDS-2017~\cite{CICIDS2017}, its WTMC-2021~\cite{engelen2021} and CRiSIS-2022~\cite{lanvin2023} refinements, and our two refinements, NFS-2023-nTE, which does not implement TCP flag expiry, and NFS-2023-TE, which does---employing the Random Forest (RF) algorithm to assess network anomaly detection in binary and multi-class classification paradigms. 
Our study builds upon previous efforts to refine these datasets, which, despite improvements, continue to exhibit anomalies such as inexplicable counts of TCP FIN and RST flags, negative values in flow features, and instances of missing data. 
These persistent inconsistencies form the backdrop against which we scrutinize the datasets, aiming to unravel how such irregularities might influence the performance and reliability of machine learning models in detecting network anomalies.

The conclusions drawn from our investigation into network anomaly are manifold. 
Firstly, the study highlights the remarkable resilience of the RF model, which consistently delivers high accuracy across all examined datasets. This consistency is a testament to the model's robustness, effectively transcending the challenges posed by data biases and measurement inaccuracies inherent in the datasets. 
Secondly, our investigation into feature importance sheds light on the nuanced role of TCP FIN and RST flags in the classification process. While these features are not consistently the most influential across all scenarios, their accurate representation proves critical in specific contexts. 
Lastly, extending our analysis to include Decision Tree (DT) and Naive Bayes (NB) algorithms, we observe a continuation of the trends identified with the RF model. This corroboration across multiple algorithms not only strengthens the validity of our findings but also underscores the broader adaptability of ML techniques to datasets with varying levels of imperfection.

Compiling these insights, the study underscores a dual reality: \textit{the encouraging adaptability of machine learning models to imperfect datasets and the potential risk of high model performance obscuring underlying data flaws}. While refined measurement techniques may not dramatically alter performance metrics, they are indispensable in ensuring that models are trained on data that truly reflects real-world network behaviors. This balance between model resilience and data integrity is crucial for advancing the field of cybersecurity, guiding future research towards the development of more accurate and reliable anomaly detection systems.

The key contributions of this paper are as follows:
\begin{enumerate}[label=(\emph{\roman*})]
    \item We provide a thorough examination of the CICIDS-2017, WTMC-2021, and CRiSIS-2022 datasets, identifying and documenting various imperfections. 
    This analysis is crucial as it brings to light specific issues that could potentially hinder the effectiveness of ML applications in network anomaly detection, thereby informing future dataset refinement and usage strategies.
    \item We introduce two refined versions of the CICIDS-2017 dataset---NFS-2023-nTE~\cite{nfs-a}, which does not implement TCP flag expiry, and NFS-2023-TE~\cite{nfs-b}, which does. 
    These refined datasets represent a step forward in terms of data integrity and reliability, and are made available online for widespread use, promoting transparency and replicability in research.
    \item We make our detailed and reproducible methodology publicly accessible as a digital artifact~\cite{repo}, which embodies the core of our contribution. This methodology underpins the findings reported in this paper and offers a versatile resource for researchers and practitioners in the field of ML-based network security. It can be readily adapted to process the CICIDS-2017 PCAP files, catering to the specific needs of diverse research endeavors. 
\end{enumerate}

In summary, the NFS-2023 datasets introduce advanced methodologies to ensure high data integrity and consistency. Unlike previous datasets, they come with comprehensive documentation and customizable measurement processes, addressing the limitations of hard-coded settings and poor reproducibility found in prior versions. Our methodology affords the flexibility to implement advanced functionalities, including TCP FIN/RST flag expiration strategies, dynamic flow labeling, and other customizable features. These capabilities enhance the measurement process, resulting in datasets that are more representative of real-world network conditions and valuable for developing robust anomaly detection models.

Our research endeavors to build upon, rather than discredit, the foundations laid by existing datasets. Ultimately, our work seeks to contribute to the evolution of machine learning in cybersecurity, aligning research efforts with the complex realities of network traffic and enhancing the resilience of systems against emerging cyber challenges.

The rest of this paper is structured as follows: 
\Cref{sec:evolution} offers a brief overview of the CICIDS-2017 dataset and its subsequent refinements, WTMC-2021 and CRiSIS-2022. 
\Cref{sec:limitations} delves into identifying the key limitations and imperfections inherent in these datasets.
\Cref{sec:objectives} outlines the primary objectives of our work, establishing the goals and motivations driving our research.
\Cref{sec:methodology} discusses the methodology employed in generating our refined versions of the CICIDS-2017 dataset. 
\Cref{sec:overview} analytically examinatnes the five datasets, focusing on aspects such as flow counts, label distributions, occurrences of negative and NaN values, and TCP FIN and RST flag counts.
\Cref{sec:results} compares the performance of the RF model across all five datasets, encompassing both binary and multi-class classification scenarios.
\Cref{sec:discussion} presents our findings and their broader implications.
\Cref{sec:limit} discusses potential future work in light of the current study's limitations.
Finally, \Cref{sec:conclusion} concludes this paper.

\section{Evolution of the CICIDS-2017 Dataset}
\label{sec:evolution}

In the realm of machine learning-based anomaly detection, the integrity and realism of datasets play a crucial role. The CICIDS-2017 dataset, in this context, has emerged as a foundational resource in the development of methodologies for network attack detection and prediction. Despite its extensive adoption in the cybersecurity community, the dataset has been subject to scrutiny due to inherent inconsistencies that could potentially skew research outcomes. This section explores the evolution of the CICIDS-2017 dataset, highlighting its original composition and subsequent refinements.


\subsection{CICIDS-2017}

The CICIDS-2017\footnote{\url{https://www.unb.ca/cic/datasets/ids-2017.html}} dataset~\cite{CICIDS2017} is a comprehensive collection of labeled network traffic patterns, encompassing benign activities and simulated malicious attacks, including DoS/DDoS, Port Scan, Brute Force, and Infiltration events. Generated over five days from 3 to 7 July 2017, this dataset offers a diverse range of flows, captured in individual PCAP files. Accompanying CSV files, produced using the CICFlowMeter\footnote{\url{https://github.com/ahlashkari/CICFlowMeter}}~\cite{icissp16,icissp17}, provide flow records with corresponding labels, facilitating the assessment and comparison of various ML-based anomaly detection approaches.



\subsection{WTMC-2021: Advancing CICIDS-2017}

Recognizing the limitations of the original CICIDS-2017 dataset, recent efforts led by \citeauthor{engelen2021} have resulted in significant improvements~\cite{engelen2021,engelen2022}. The WTMC-2021 dataset\footnote{\url{https://intrusion-detection.distrinet-research.be/WTMC2021/tools_datasets.html}}, a refined iteration, addresses key issues in traffic generation, flow construction, feature extraction, and labeling identified in the original dataset. This refined version, which revises over 20 percent of the original traffic data, offers a more accurate and reliable resource for anomaly detection research, setting a new standard in dataset integrity for the cybersecurity community.


\subsection{CRiSIS-2022: Further Refinement}

Building upon the advancements of WTMC-2021, the CRiSIS-2022 dataset\footnote{\url{https://gitlab.inria.fr/mlanvin/crisis2022}} introduced by \citeauthor{lanvin2023} offers additional enhancements~\cite{lanvin2023}. This version addresses overlooked issues such as mislabeled port scan attacks and duplicated traffic captures. By rectifying these aspects, CRiSIS-2022 provides a more robust and accurate dataset, further aligning simulated network traffic conditions with real-world scenarios and bolstering the reliability of subsequent cybersecurity research.



\section{Key Limitations in Dataset Realism and Applicability}
\label{sec:limitations}


Despite the release of the refined CICIDS-2017 and the remedial patches applied to the CICFlowMeter tool, the application of this dataset for research transferable to real-world scenarios remains limited. The CICFlowMeter tool has been the cornerstone for all existing versions of the CIC-IDS-2017 dataset, including the original and the improved ones. Although patches have been implemented to rectify the tool, its flow construction and feature extraction pipeline still diverge from the operations followed in real-world flow metering. The limitations can be grouped into two primary categories.

\subsection{Inadequate Implementation of Flow Timeout Mechanism}

The CICFlowMeter tool, diverges from standard flow metering practices in its implementation of flow timeouts. While it employs a flow timeout akin to an idle timeout, it lacks a standard active timeout mechanism. Instead, it uses an activity timeout for subflow statistics, differing from the de facto standard in flow metering tools and protocols like NetFlow and IPFIX. This absence of a conventional flow expiration policy results in datasets that are not fully representative of typical network conditions, a limitation present in both the original and patched versions of CICFlowMeter.

\subsection{Flaws in TCP Flow Flag-Based Expiration Mechanism}

The original CICFlowMeter was claimed lacking a robust TCP flag-based flow expiration mechanism, terminating TCP flows upon the first observation of FIN or RST flags. This was addressed by the patches introduced by \citeauthor{engelen2021}, which aligned flow termination more closely with the TCP specification. In the patched version, a TCP flow is no longer terminated after a single FIN packet. It terminates after mutual exchange of FIN packets, which is more in line with the TCP specification. Additionally, an RST packet is no longer ignored either. Instead, the RST packet also terminates a TCP flow. 

However, this modified termination process, while technically accurate, deviates from the behavior commonly observed in commercial flow metering tools, which typically terminate flows upon the first FIN or RST flag. This discrepancy introduces potential biases in performance assessments, as the datasets do not accurately reflect real-world flow metering behavior.

Additionally, and more importantly, with the patch, a critical issue was introduced. The WTMC-2021 and CRiSIS-2022 datasets exhibit anomalously high FIN and RST flag counts in individual flow records, with counts sometimes exceeding 100. Such behavior is atypical in standard flow metering practices, leading to ambigous interpretability and relevance for real-world applications.




\begin{table*}[ht]
\centering
\caption{Summary of PCAP Files Analysis}
\label{tbl:pcap-analysis}
\setlength\tabcolsep{2pt} 
\begin{tabular*}{\tblwidth}{@{}L|R|R|R|R|R|R@{}}
\toprule
\textbf{PCAP} & \textbf{No. of Packets} & \textbf{No. of Duplicates} & \textbf{\% Duplicates} & \textbf{No. of Frames} & \textbf{No. of Out of Order Frames} & \textbf{\% Out of Order} \\ \midrule
Monday    & 11 709 971 & 526 729 & -    & 11 183 242 & 3 105 & -     \\ 
Tuesday   & 11 551 954 & 496 702 & 4.3\% & 11 055 252 & 3 581 & 0.03\% \\ 
Wednesday & 13 788 878 & 494 582 & 3.6\% & 13 294 296 & 12 531 & 0.09\% \\
Thursday  & 9 322 025  & 568 505 & 6.1\% & 8 753 520  & 3 424 & 0.04\% \\
Friday    & 9 997 874  & 480 540 & 4.8\% & 9 517 334  & 6 980 & 0.07\% \\
\bottomrule
\end{tabular*}
\end{table*}


\section{Objectives}
\label{sec:objectives}

Our research is guided by three central objectives.

\subsection{Addressing Identified Issues with a Methodologically Sound Dataset}

The primary goal of our research was to address the critical issues highlighted in \Cref{sec:limitations} by developing a dataset through a process that is both transparent and methodologically robust. The CICFlowMeter, widely recognized in AI and ML-based anomaly detection research, suffers from limited documentation, scalability issues, methodological errors, and constraints in customizability. Its open-source nature does not mitigate the challenge of deciphering and resolving its complex underlying issues. Moreover, although there is a hint about an impending Python-based version that might address these issues\footnote{\url{https://github.com/ahlashkari/CICFlowMeter/issues/154}}, such a release had not materialized as of the date of our study.

In our search for a tool with methodologically sound flow feature calculation, we utilized NFStream, offering dependable performance and flexibility in customization. It is particularly adept at implementing TCP flow expiration policies and advanced flow labeling methodologies. By utilizing NFStream, we were able to sidestep the intricate challenges posed by CICFlowMeter, effectively addressing the limitations highlighted in this present work. As a result, we generated two novel versions of the CICIDS-2017 dataset:
\begin{itemize}
\item \textit{NFS-2023-nTE}: This version disabled TCP flag-based flow expiration, aiming to mirror the flow generation process used in existing dataset versions.
\item \textit{NFS-2023-TE}: In contrast, this version enabled TCP flag-based flow expiration, better reflecting the flow characteristics observed in real-world network flow metering scenarios.
\end{itemize}

\subsection{Comparing ML Model Performances Across Datasets}

Our second objective focused on conducting a comparative analysis of ML model performances across several datasets. This encompassed the original CICIDS-2017 dataset, along with its WMTC-2021 and CRiSIS-2022 refinements, and the two novel versions we created using NFStream: NFS-2023-nTE (without TCP flag expiry) and NFS-2023-TE (with TCP flag expiry). The purpose of this comparison was to evaluate the influence of different flow metering strategies on ML model outcomes. A key aspect of this analysis was to understand how the incorporation or absence of TCP flag expiration policies affects the accuracy and reliability of these models in network traffic analysis and anomaly detection.



\subsection{Developing a Customizable Dataset Creation Methodology}

Finally, our present work was also aimed at developing a methodology that researchers can customize to process PCAP files as per their unique requirements. This approach is designed to empower the research community to create their own versions of datasets, grounded in a sound, transparent, and adaptable data processing methodology.

\section{Methodology}
\label{sec:methodology}

In this sections, we delve into the specifics of our methodology, discussing the rationale behind our configuration choices and the techniques employed to ensure the integrity, accuracy, and relevance of our datasets and achieved results.

\subsection{Raw Data Preprocessing}

Our experimental approach began with downloading the raw packet trace files from the CICIDS-2017 dataset. In line with the preprocessing steps outlined by \citeauthor{lanvin2023}~\cite{lanvin2023}, we undertook a similar strategy to minimize the undocumented (likely) anomalies in the raw packet traces~\cite{lanvin2023} that could potentially bias our results.

Initially, we utilized the editcap command (\texttt{editcap -D 10000 input.pcap output.pcap}) to remove duplicate packets from the acquired PCAP files. This command, with a deduplication window set to 10,000 packets, effectively identifies and eliminates duplicates, striking a balance between computational efficiency and the thoroughness required for our analysis.

Following deduplication, the reordercap command (\texttt{reordercap input.pcap output.pcap}) was applied to the cleaned traffic trace files. The purpose of this step was to reorder the packets chronologically, thereby ensuring the temporal accuracy of network events, which is crucial for a precise analysis of network flow.

\Cref{tbl:pcap-analysis} provides an in-depth summary of the results obtained from executing these commands on each raw packet trace file. It details the total count of frames in the PCAP files before and after deduplication and highlights the number of frames that were out of sequence pre- and post-reorder, along with their respective percentage differences. Through these preprocessing steps, we were able to refine the raw PCAP files effectively, laying a solid foundation for our subsequent network traffic flow analysis.

\subsection{IP Flow Generation}

For the critical task of processing the PCAP files, we employed NFStream~\cite{AOUINI2022108719}, our Python-based tool that is specifically engineered for rapid, flexible, and expressive data handling in network analysis.

The core strength of NFStream lies in its hybrid design, which combines the ease and accessibility of Python with the performance efficiency of C. This is epitomized by the NFlow structure, a central component written in C, that forms the backbone of the tool. The NFlow structure is designed to represent a network flow and incorporates essential attributes like the five-tuple flow key. Beyond these fundamental features, it is augmented with a rich set of flow statistics and metadata, encompassing categories such as core features, Layer 7 visibility, post-mortem statistics, and Sequence of Packet Length and Time (SPLT) details. What makes NFStream particularly versatile is the ability for users to selectively enable or disable these feature sets during the flow measurement process, thereby tailoring the analysis to specific requirements.

Moreover, NFStream excels in bridging the gap between raw network measurements and sophisticated data science analytics. Its intrinsic capabilities in flow measurement and feature computation are complemented by an adaptable architecture. This flexibility is particularly pronounced in the integration of bespoke network functionalities through the NFPlugin component~\cite{AOUINI2022108719}. In our study, this feature was instrumental, as it allowed us to integrate our specific TCP flow expiration policy and flow labeling methodology directly into the tool. 

\subsubsection{Flow Measurement Specificities}

In our research, we utilized NFStream v6.5.4 and primarily adhered to the default settings recommended in the NFStream documentation. However, to align the tool's functionality with the specific requirements of our dataset and research objectives, we customized several parameters as follows:

\begin{itemize}
    \item \textit{Decode Tunnels:} We disabled the tunnel decoding feature because the CICIDS-2017 PCAP files did not contain any tunneled traffic. This deactivation ensured a streamlined analysis focused on the traffic types present in our dataset.
    \item \textit{BPF Filter:} A custom Berkeley Packet Filter (BPF) was implemented to exclusively capture TCP and UDP packets over IPv4. With this customization, we achieved a more targeted packet filtering process, effectively excluding traffic types of marginal volume.
    \item \textit{Flow Timeouts:} We adjusted the idle and active timeouts to 60 and 120 seconds, respectively. This change from the default settings (120 and 1800 seconds) was driven by the aim to more accurately replicate the network traffic flow metering conditions in CICIDS-2017, thereby enhancing the realism, relevance and comparability of our findings.
    \item \textit{DPI-based Labeling:} We set the `n\_dissections' value to~0, effectively disabling Deep Packet Inspection (DPI)-based Layer 7 labeling. This adjustment was made considering its limited utility in our Intrusion Detection System (IDS) analysis.
    \item \textit{Statistical Analysis:} In deviation from the default configuration, we enabled the statistical analysis feature. This decision was made to enrich our dataset with detailed flow statistics, providing a more robust foundation for comprehensive data evaluation.
\end{itemize}


\subsubsection{TCP Flow Expiration Policy}
\label{sec:TCP-policy}

Our TCP flow expiration policy, implemented as an NFPlugin within NFStream, uses a specific set of criteria for marking TCP flows as expired based on the packet characteristics. This policy is crucial for ensuring accurate dataset generation and flow management.

\begin{itemize}
    \item \textit{At flow initiation:} Upon the initiation of a new network flow, our policy dictates that if the incoming packet carries a TCP RST flag (indicating an abrupt termination) or a TCP FIN flag (signifying a normal closure), the flow is immediately marked for expiration using a specific identifier. Specifically, the expiration ID of `-1` is assigned, which is reserved for user-defined expiration in NFStream. This ensures that our dataset accurately reflects both abrupt terminations and orderly closures of TCP connections right from the start.
    \item \textit{At flow update:} For existing network flows, our policy similarly mandates the assignment of the expiration ID `-1` if a packet carries an RST flag or if a FIN flag is detected during flow updates. This practice ensures that changes in the status of ongoing connections, whether unexpected interruptions or planned closures, are promptly captured.
\end{itemize}

The choice of `-1` for the expiration ID allows NFStream to distinguish these flows as requiring immediate removal from the flow cache, which maintains records of active flows. This is in contrast to other values like `0` for idle timeout and `1` for active timeout, which NFStream uses to manage flows under standard operational conditions.

\begin{algorithm}[!ht]
\caption{Flow Expiration Manager for TCP Flows}
\label{al:a1}
\begin{algorithmic}[1]

\State \textbf{class} FlowExpirationManager \textbf{inherits} NFPlugin:
\State \Comment{\textcolor{gray}{Manages the expiration policy for TCP flows.}}

\Procedure{on\_init}{$packet, flow$}
    \State \Comment{\textcolor{gray}{Set the expiration ID based on RST or FIN flags.}}
    \If{$packet.rst \ \textbf{or} \ packet.fin$}
        \State $flow.expiration\_id \gets -1$
    \EndIf
\EndProcedure

\Procedure{on\_update}{$packet, flow$}
    \State \Comment{\textcolor{gray}{Set the expiration ID based on RST or FIN flags.}}
    \If{$packet.rst \ \textbf{or} \ packet.fin$}
        \State $flow.expiration\_id \gets -1$
    \EndIf
\EndProcedure

\end{algorithmic}
\end{algorithm}

\Cref{al:a1} shows the pseudo code for our TCP flow expiration strategy within NFStream.
This strategy, while not encompassing every nuanced TCP sequence anomaly present in raw traffic traces, is in line with established methodologies used by various flow exporters, both open-source and commercial, as referenced in literature~\cite{6814316}. This includes widely-used protocols like IPFIX and NetFlow, which are integral to numerous vendor devices. This alignment with common industry practices aims to boost the practical applicability of our datasets.

\subsubsection{Post-Processing Flows}
\label{sec:post-processing}

To further enhance the relevance and accuracy of our dataset, we implement an additional refinement step after generating the flow data. This step involves filtering out flow records that may not provide meaningful insights due to the nature of their termination. Specifically, we target flows that were terminated prematurely by an RST or FIN flag in their initial packet, which could represent truncated connections or abnormal operations.

\Cref{al:a2} outlines the specific criteria used to filter out certain flow records during our post-processing phase. This process helps to ensure that our dataset focuses on more substantive network connections that offer deeper and more contextually rich insights for analysis. 
Such filters are particularly important in studies of anomaly detection, as they help to eliminate potential noise and focus on genuinely anomalous activities that are not artifacts of flow expiration policies.

\begin{algorithm}[!h]
\caption{Post-Processing Flows}
\label{al:a2}
\begin{algorithmic}[1]

\Procedure{FilterFlows}{$flows$}
    \State \Comment{\textcolor{gray}{Exclude flows that may skew analysis results.}}
    \State Initialize $filtered\_flows$
    \For{each $row$ in $flows$}
        \If{$row[bidirectional\_packets] \neq 1$
        \textbf{and} ($row[bidirectional\_rst\_packets] \neq 1$
        \textbf{or} $row[bidirectional\_fin\_packets] \neq 1$)}
            \State Add $row$ to $filtered\_flows$
        \EndIf
    \EndFor
\EndProcedure

\end{algorithmic}
\end{algorithm}

\subsubsection{Attack Type Labelling}

In our study, we employed an additional NFPlugin within the NFStream framework to systematically label each network flow based on predefined criteria. These criteria are designed to reflect the attack patterns and benign behaviors documented for each weekday from Monday to Friday in the original dataset~\cite{icissp17}. We analyze key flow characteristics such as source and destination IPs, ports, protocols, and packet payloads. Each characteristic is matched against specific attack profiles that include not only the types of attacks but also their known time windows through a detailed multi-stage process:
\begin{enumerate}
    \item \textit{Attribute Matching:} Each flow’s attributes are first checked against the attack profiles to identify potential matches based on IP addresses, port numbers, and protocols.
    \item \textit{Temporal Validation:} For flows that match the attack profiles in attributes, the time stamps are then verified against the documented time windows of the attacks to confirm if the flow occurred during the known period of attack activity.
    \item \textit{Final Labeling:} Flows that match both attribute and temporal criteria are labeled with specific attack tags (e.g., `SSH-Patator', `Web Attack - Brute Force'), whereas those that do not align with any known attack patterns are marked as `BENIGN'.
\end{enumerate}

An integral part of our labeling process is the management of flows with zero packet payloads (ZPL). These are particularly challenging as they can skew the accuracy of anomaly detection. Our payload manager maintains a detailed record of the flow payloads, enabling us to identify and appropriately label ZPL flows as `BENIGN'. This adjustment is necessary because, despite being marked as anomalies in the original dataset, such flows might not exhibit typical malicious characteristics~\cite{engelen2021}. In our methodology, we label flows with ZPL as `BENIGN', except for those identified as part of a `Portscan' type.

Additionally, our labeling mechanism includes the capability to reverse the flow direction. This feature is essential for accurately labeling flows in scenarios where packets from the same flow may be segmented into subflows (as a consequnce of our TCP flow expiration strategy). By swapping source and destination parameters, we ensure that each part of the flow is evaluated in its proper context, enhancing the overall accuracy of our dataset labeling.

This meticulous approach ensures a high degree of accuracy in distinguishing between benign and malicious flows, reflecting the unique attack dynamics documented for each day. This systematic process facilitates more targeted and meaningful analysis in subsequent research phases, enabling robust detection of anomalies and attack patterns.

\section{Dataset Overview}
\label{sec:overview}

\subsection{Distribution of Benign and Anomalous Flows}

\Cref{tbl:datasets} presents a comparative analysis of the distribution of the flow types across the five CICIDS-2017 dataset versions examined in this study.

The CICIDS-2017 dataset is available in two variants. The version tailored for ML research omits several columns, including Flow ID, Source IP, Source Port, Destination IP, Protocol, and Timestamp, unlike its counterpart which includes these essential fields. Given the significance of the protocol identifier in model training, we opted for the more comprehensive version, despite its inclusion of 288 602 flows with unknown labels. These unlabeled flows were excluded from our analysis.

For the WTMC-2021 dataset, we disregarded flows marked as `Attempted'. As noted by \citeauthor{engelen2021}~\cite{engelen2021}, some flows in payload-reliant attack categories lack an actual payload and are thus labeled as `Attempted'.

From the data in \Cref{tbl:datasets}, it is evident that the CICIDS-2017 and NFS-2023-TE datasets exhibit a roughly similar count and distribution of flow records. These datasets share a common approach in expiring flows upon encountering the first FIN or RST TCP flag. Conversely, the WTMC-2021, CRiSIS-2022, and NFS-2023-nTE datasets, which were generated without TCP FIN or RST flag-based flow expiration, demonstrate a closely aligned flow count and distribution pattern.

A noteworthy distinction emerged regarding the integrity of data fields in the datasets. All datasets produced using CICFlowMeter, exhibited a small but significant number of flow entries with NaN (Not a Number) values. This occurrence of NaN values potentially indicates gaps or inconsistencies in the flow metering methodology employed by CICFlowMeter. Such anomalies can have a cascading effect on data quality and, consequently, on the performance and reliability of ML models trained using these datasets. In contrast, the datasets generated using NFStream demonstrated a higher level of data completeness, with no instances of unfilled data fields. 

In addition to the occurrence of NaN values, our analysis uncovered another dimension of data integrity concerns in the datasets produced using CICFlowMeter. A significant number of entries across various features, such as `Flow Duration', `Flow Bytes/s', `Flow Packets/s', and several others, contained negative values for each day from Monday to Friday. Such negative values are not logically justifiable within the context of these network traffic flow metrics and suggest the presence of errors in the dataset's measurement process. This observation further underscores the challenges associated with ensuring data quality in the datasets derived from CICFlowMeter.


Conversely, while the WTMC-2021 dataset shows an improvement with fewer features containing negative numbers, it still presents this anomaly, particularly from Monday to Thursday in the `Flow IAT Min' feature and more extensively on Friday. This indicates that although refinements have been made, certain measurement inaccuracies persist. The CRiSIS-2022 dataset stands out for its absence of such irregularities, indicating a more rigorous flow measurement methodology or post-processing where such flow entries might have been removed manually. 

While the NFStream-generated datasets do contain entries with negative numbers, these are intentionally designed to represent specific flow conditions, such as custom flow expiration policies, rather than being indicative of measurement errors. The negative values in NFStream datasets are, therefore, a feature of the dataset's structure and not a flaw.

Nonetheless, this comparison underscores the influence of flow expiration policies on the structure and composition of network traffic datasets, thereby potentially affecting the subsequent analysis and model training processes.

\begin{table}[!htbp]
    \scriptsize
    \centering
    \begin{threeparttable}
    \caption{Flow type distribution overview per dataset version}
    \label{tbl:datasets}
    \setlength\tabcolsep{3.4pt} 
    \begin{tabular*}{\tblwidth}{@{}C|L|RRR|R|R@{}}
    \toprule
      \rotatebox[origin=c]{90}{\textbf{Day}} 
      & \multicolumn{1}{c|}{\textbf{Dataset}} 
      & \multicolumn{1}{c}{\textbf{Total}} 
      & \multicolumn{1}{c}{\textbf{Benign}} 
      & \multicolumn{1}{c|}{\textbf{Anomaly}} 
      & \multicolumn{1}{c|}{\textbf{NaN}}
      & \multicolumn{1}{c}{\textbf{Negative}}
      \\
    \midrule
        \multirow{5}{*}{\rotatebox[origin=c]{90}{\textbf{Mon}}}
        & CICIDS-2017 & {529 918} & {529 918} & \multicolumn{1}{c|}{\multirow{5}{*}{---}} & 64 & 299 597\\
        & WTMC-2021 & {371 749} & {371 749} & & 47 & 517 \\
        & CRiSIS-2022 & {372 425} & {372 425} & & 2 492 & 0\\
        \cmidrule[0.5pt]{2-4} \cmidrule[0.5pt]{6-7}
        & NFS-2023-nTE & {376 826} & {376 826} & & 0 & \ddag \\
        & NFS-2023-TE & {559 775} & {559 775} & & 0 & \ddag \\
    \midrule
        \multirow{5}{*}{\rotatebox[origin=c]{90}{\textbf{Tue}}} 
        & CICIDS-2017 & 445 909 & 432 074 & 13 835 & 201 & 257 104 \\
        & WTMC-2021\hspace{1.45mm}\dag & 321 984 & 315 031 & 6 953  & 23 & 511 \\
        & CRiSIS-2022 &  322 462 & 315 509 & 6 953  & 2 336 & 0 \\
        \cmidrule[0.5pt]{2-7}
        & NFS-2023-nTE & 325 677 & 318 705 & 6 972  & 0 & \ddag \\
        & NFS-2023-TE & 467 254 & 456 311 & 10 943  & 0 & \ddag \\
    \midrule
        \multirow{5}{*}{\rotatebox[origin=c]{90}{\textbf{Wed}}} 
        & CICIDS-2017 & 692 703 & 440 031 & 252 672 & 1 008 & 336 224\\
        & WTMC-2021\hspace{1.45mm}\dag & 491 006 & 319 216 & 171 790 & 29 & 758 \\
        & CRiSIS-2022 & 497 115 & 325 324 & 171 791 & 2 281 & 0 \\
        \cmidrule[0.5pt]{2-7}
        & NFS-2023-nTE & 503 140 & 329 267 & 173 873  & 0 & \ddag \\
        & NFS-2023-TE & 865 413 & 691 015 & 174 398  & 0 & \ddag \\
    \midrule
        \multirow{5}{*}{\rotatebox[origin=c]{90}{\textbf{Thu}}} 
        & CICIDS-2017 * & 458 968 & 456 752 & 2 216  & 38 & 267 336 \\
        & WTMC-2021\hspace{1.45mm}\dag & 360 486 & 360 264 & 222    & 288 & 621 \\
        & CRiSIS-2022 & 355 618 & 291 211 & 64 407 & 33 287 & 0\\
        \cmidrule[0.5pt]{2-7}
        & NFS-2023-nTE & 358 302 & 294 128 & 64 174  & 0 & \ddag \\
        & NFS-2023-TE & 485 224 & 413 744 & 71 480  & 0 & \ddag \\
    \midrule
        \multirow{5}{*}{\rotatebox[origin=c]{90}{\textbf{Fri}}} 
        & CICIDS-2017 & 703 245 & 414 322 & 288 923 & 47 & 284 166 \\
        & WTMC-2021\hspace{1.45mm}\dag & 546 445 & 291 433 & 255 012 & 348 & 515 \\
        & CRiSIS-2022 & 548 828 & 293 367 & 255 461 & 3 011 & 0 \\
        \cmidrule[0.5pt]{2-7}
        & NFS-2023-nTE & 547 186 & 293 341 & 253 845 & 0 & \ddag \\
        & NFS-2023-TE & 775 694 & 518 961 & 256 733 & 0 & \ddag \\
    \bottomrule
    \end{tabular*}
    \begin{tablenotes}
        \item[*] Contains 288 602 flows with \textit{NaN} labels whose count is not included.
        \item[\dag] Contains attempted attack types whose count is not included.
        \item[\ddag] Contains negative numbers but these denote specific flow conditions. 
    \end{tablenotes}
    \end{threeparttable}
\end{table}

\subsection{Distribution of Specific Anomaly Types}

Building on the insights from \Cref{tbl:datasets}, \Cref{tbl:anomalies} delves deeper into the distribution of specific anomaly types across the five versions of the CICIDS-2017 dataset. The data presented in this table excludes the NaN entries detailed in \Cref{tbl:datasets}.

\begin{table*}[!ht]
\centering
\caption{Anomaly breakdown}
\label{tbl:anomalies}
\begin{tabular*}{\tblwidth}{@{}L|L|R|R|R|R|R@{}}
\toprule
\textbf{DS} & Anomaly Type & CICIDS-2017 & WTMC-2021 & CRiSIS-2022 & NFS-2023-nTE & NFS-2023-TE \\ 
    \midrule
        \textbf{Mon} & --- & --- & --- & --- & --- & --- \\ 
    \midrule
        \multirow{2}{*}{\textbf{Tue}} 
            & FTP-Patator & 7 938 & 3 973 & 3 973 & 3 992 & 7 963\\ 
            & SSH-Patator & 5 897 & 2 980 & 2 980 & 2 980 & 2 980\\ 
    \midrule
        \multirow{5}{*}{\textbf{Wed}} 
            & DoS GoldenEye & 10 293 & 7 567 & 7 567 & 7 916 & 7 917 \\ 
            & DoS Hulk & 231 073 & 158 469 & 158 470 & 158 027 & 158 546 \\ 
            & DoS Slowhttptest & 5 499 & 1 742 & 1 742 & 2 727 & 2 732 \\
            & DoS Slowloris & 5 796 & 4 001 & 4 001 & 5 192 & 5 192 \\ 
            & Heartbleed & 11 & 11 & 11 & 11 & 11 \\ 
    \midrule
        \multirow{5}{*}{\textbf{Thu}} 
            & Infiltration & 36 & 32 & 32 & 27 & 28 \\             
            & PortScan & -- & -- & 64 185 & 63 957 & 71 262 \\ 
            & Web Attack - Brute Force & 1 507 & 151 & 151 & 151 & 151 \\ 
            & Web Attack - SQL Injection & 21 & 12 & 12 & 12 & 12 \\ 
            & Web Attack - XSS & 652 & 27 & 27 & 27 & 27 \\ 
    \midrule
        \multirow{3}{*}{\textbf{Fri}} 
            & Bot & 1 966 & 738 & 738 & 738 & 738 \\ 
            & DDoS & 128 027 & 95 123 & 95 144 & 93 178 & 95 685\\   
            & PortScan & 158 930 & 159 151 & 159 579 & 159 929 & 160 310\\ 
\bottomrule
\end{tabular*}
\end{table*}

\begin{table*}[!ht]
    \centering
    \caption{Number of flows more than two with FIN and RST flags and their distribution between benign and anomalous traffic types}
    \label{tbl:fin-rst}
    \begin{tabular*}{\tblwidth}{@{}C|L|RRR|RRR@{}}
    \toprule
      \textbf{Day} & \multicolumn{1}{c|}{\textbf{Dataset}} 
      & \multicolumn{3}{c|}{\textbf{Flows with FIN $>2$}} 
      & \multicolumn{3}{c}{\textbf{Flows with RST $>2$}} \\
      & & \textbf{Total} & \textbf{Benign} & \textbf{Anomaly} 
      & \textbf{Total} & \textbf{Benign} & \textbf{Anomaly} \\ 
    \midrule
       \multirow{5}{*}{\textbf{Mon}}
        & CICIDS-2017 & 0 & 0 & 0 & 0 & 0 & 0 \\
        & WTMC-2021   & 18 703 & 18 703 & 0 & 9 860 & 9 860 & 0  \\
        & CRiSIS-2022 & 16 477 & 16 477 & 0 & 9 657 & 9 657 & 0 \\
        \cmidrule[0.5pt]{2-8}
        & NFS-2023-nTE  & 16 476 & 16 476 & 0 & 9 785 & 9 785 & 0 \\
        & NFS-2023-TE  & 0 & 0 & 0 & 0 & 0 & 0 \\
    \midrule
       \multirow{5}{*}{\textbf{Tue}}
        & CICIDS-2017 & 0 & 0 & 0 & 0 & 0 & 0 \\
        & WTMC-2021   & 11 737 & 11 690 & 47 & 4 396 & 4 383 & 13  \\
        & CRiSIS-2022 & 9 387 & 9 340 & 47 & 4 152 & 4 139 & 13 \\
        \cmidrule[0.5pt]{2-8}
        & NFS-2023-nTE  & 9 386 & 9 339 & 47 & 4 217 & 4 204 & 13 \\
        & NFS-2023-TE  & 0 & 0 & 0 & 0 & 0 & 0 \\
    \midrule
       \multirow{5}{*}{\textbf{Wed}}
        & CICIDS-2017 & 0 & 0 & 0 & 0 & 0 & 0 \\
        & WTMC-2021   & 10 952 & 5 692 & 5 260 & 40 779 & 2 512 & 38 267  \\
        & CRiSIS-2022 & 8 566 & 3 305 & 5 261 & 40 577 & 2 309 & 38 268 \\
        \cmidrule[0.5pt]{2-8}
        & NFS-2023-nTE  & 8 827 & 3 295 & 5 532 & 40 832 & 2 380 & 38 452 \\
        & NFS-2023-TE  & 0 & 0 & 0 & 0 & 0 & 0 \\
    \midrule
       \multirow{5}{*}{\textbf{Thu}}
        & CICIDS-2017 & 0 & 0 & 0 & 0 & 0 & 0 \\
        & WTMC-2021   & 5 272 & 5 272 & 0 & 2 076 & 2 076 & 0  \\
        & CRiSIS-2022 & 2 231 & 2 231 & 0 & 1 884 & 1 884 & 0 \\
        \cmidrule[0.5pt]{2-8}
        & NFS-2023-nTE  & 2 230 & 2 230 & 0 & 2 028 & 2 028 & 0 \\
        & NFS-2023-TE  & 0 & 0 & 0 & 0 & 0 & 0 \\
    \midrule
       \multirow{5}{*}{\textbf{Fri}}
        & CICIDS-2017 & 0 & 0 & 0 & 0 & 0 & 0 \\
        & WTMC-2021   & 4 877 & 4 826 & 46 & 2 434 & 2 425 & 9  \\
        & CRiSIS-2022 & 2 624 & 2 578 & 46 & 2 210 & 2 201 & 9 \\
        \cmidrule[0.5pt]{2-8}
        & NFS-2023-nTE  & 3 735 & 2 554 & 1 181 & 3 621 & 3 608 & 13 \\
        & NFS-2023-TE  & 0 & 0 & 0 & 0 & 0 & 0 \\
    \bottomrule
    \end{tabular*}
\end{table*}


An important observation from \Cref{tbl:anomalies} is the alignment in value counts between the two refined versions and those measured using NFStream, when compared to the original CICIDS-2017 dataset. This alignment suggests that the updates and patches applied to CICFlowMeter have likely contributed to a more accurate measurement and labeling of anomalies, bringing these metrics closer to the results observed with NFStream.

Another notable observation from \Cref{tbl:anomalies} is the effective identification and labeling of PortScan attacks in the CRiSIS and NFStream versions of the dataset. These attacks were previously unnoticed in the original CICIDS-2017 dataset and its WMTC-2021 refinement. The enhanced accuracy in measuring and labelling PortScan is a critical advancement for network security research, thereby offering a more robust foundation for training and evaluating ML models in the context of anomaly detection.

\subsection{Reconstructing TCP FIN and RST Flag Counts}

\Cref{tbl:fin-rst} provides an insightful comparison of the counts of flows with three or more TCP FIN and RST flags across the five studied versions of the CICIDS-2017 dataset. This analysis is crucial for verifying the impact of TCP flag-based flow expiration policies on the characteristics of the generated flow records.

The enhancement of TCP flag-based flow expiration in WMTC-2021 was intended to align flow records more closely with standard TCP operational behaviors. Ideally, this would result in flows being expired after a bidirectional exchange of packets with FIN flags, or immediately upon detecting a packet with an RST flag. However, our analysis reveals that the CICFlowMeter version used for WMTC-2021 did not fully achieve this objective. Contrary to expectations, both the WMTC-2021 and CRiSIS-2022 datasets include flow records with anomalously high counts of FIN or RST flags, sometimes exceeding 100 flags per a single flow in extreme cases.

Our preliminary examination of the CICFlowMeter's patched version revealed that the abnormally high TCP flag counts are likely due to how flow expiration is handled. In the patched CICFlowMeter, TCP flows are designated for expiration upon detecting either a TCP RST flag or a sequence of two TCP FIN flags followed by a TCP ACK flag. However, even after a flow is marked for expiration, its attributes continue to be updated until it is eventually removed from the flow cache. This removal occurs either due to the active timeout (hard coded 120 seconds) being exceeded or when a TCP SYN flag is encountered in a flow already marked for expiration. In the latter case, a new flow record is initiated with the packet containing the SYN flag, and the previous flow is discarded. The result is flow records with significantly high counts of these flags, making their interpretability ambiguous and, at times, misleading with respect to standard TCP operation.

Our measurement with NFStream revealed differences in dataset characteristics based on whether TCP compliant expiration policy was enabled or disabled. When we activated the TCP expiration policy in NFStream, the resulting dataset (NFS-2023-nTE) aligned well with the expected standard TCP behavior. The flows were terminated in a manner consistent with typical TCP connection closures, leading to a more realistic representation of network traffic. However, this version diverged from the flow characteristics observed in the WMTC-2021 and CRiSIS-2022 datasets, particularly in terms of the counts of TCP FIN and RST flags.

Conversely, when the TCP expiration policy was disabled, the resulting dataset (NFS-2023-nTE) showed similarities to the flow characteristics observed in WMTC-2021 and CRiSIS-2022, as evident from \Cref{tbl:fin-rst}. Given this alignment, we chose to produce the NFS-2023-nTE dataset without enabling the TCP flag-based expiration policy. This decision was pivotal in ensuring that the results obtained using NFS-2023-nTE could be more effectively contrasted with those derived from existing datasets.

While not directly evident from \Cref{tbl:fin-rst}, in the NFS-2023-TE dataset---which was generated without activating the TCP flag-based expiration policy---there were no flows with more than two counts of TCP FIN or RST flags. In fact, these counts were limited to a maximum of one per flow, perfectly aligning with the expected behavior outlined in our TCP expiration policy described in \Cref{sec:TCP-policy}. 

Nonetheless, the discrepancy between the expected and observed behaviors in flow records with respect to TCP FIN and RST flags underscores the importance of accurate implementation of network protocols in flow metering tools. It also highlights the effectiveness of NFStream in adhering to TCP standards when the appropriate expiration policy is activated.




\section{ML Performance Comparison}
\label{sec:results}



\subsection{ML Algorithm}

In our study, we explore the impact of data integrity on network anomaly detection by employing Random Forest (RF), a widely recognized and frequently used machine learning technique. RF is an ensemble learning method that operates by constructing a multitude of decision trees at training time. It outputs the mode of the classes (classification) of the individual trees, which enhances the overall predictive accuracy and controls over-fitting.

The rationale behind this selection is that RF have been consistently utilized in the literature related to the CICIDS-2017, WMTC-2021, and CRiSIS-2022 datasets. For instance, CICIDS-2017 employed the K-Nearest Neighbors (KNN), RF, Iterative Dichotomiser 3 (ID3), Adaboost, Multilayer Perceptron (MLP), NB, and Quadratic Discriminant Analysis (QDA) algorithms~\cite{icissp17}; WMTC-2021 focused solely on RF~\cite{engelen2021}; and CRiSIS-2022 assessed Support Vector Machine (SVM), NB, RF, and DT~\cite{lanvin2023}. 
Choosing the RF algorithm ensures that we evaluate a method previously employed in related research, thereby enhancing the comparability and transferability of our observations across these studies.

\subsection{Feature Selection}

In our comparative study of RF model performances using different versions of the CICIDS-2017 dataset, a crucial step is the mapping of flow features generated by CICFlowMeter and NFStream. While both tools serve similar purposes in network traffic analysis, they generate a range of flow features with some overlaps and some unique aspects\footnote{For a comprehensive list of features from CICFlowMeter, see\\ \url{https://github.com/ahlashkari/CICFlowMeter/blob/master/ReadMe.txt}}\textsuperscript{,}\footnote{For NFStream features, refer to \url{https://www.nfstream.org/docs/api}}. Our focus is on identifying features that are common across both tools to ensure a fair and balanced comparison. 
\Cref{tbl:feature-mapping} details this mapping, highlighting the five-tuple (five flow key attributes) and 41 flow features that can be directly compared or have closely analogous counterparts. By concentrating on a consistent set of common flow features, we aim to provide a logical and well-grounded basis for comparing the effectiveness of methodologies developed using the datasets derived from these two tools.

\begin{table}[!ht]
\footnotesize
\centering
\begin{threeparttable}
\caption{Flow feature mapping between NFStream and CICFlowMeter}
\label{tbl:feature-mapping}
\begin{tabular*}{\tblwidth}{@{}L|L@{}}
\toprule
\textbf{NFStream Feature}                 & \textbf{CICFlowMeter Feature}         \\ 
\midrule
src\_ip                                   & Source IP \dag                              \\ 
src\_port                                 & Source Port  \dag                            \\ 
dst\_ip                                   & Destination IP  \dag                              \\ 
dst\_port                                 & Destination Port \dag                             \\ 
protocol                                  & Protocol                              \\ 
\midrule
src2dst\_packets                          & Total Fwd Packet  \dag                    \\ 
dst2src\_packets                          & Total Bwd packets  \dag                   \\ 
src2dst\_bytes                            & Total Length of Fwd Packet \dag           \\ 
dst2src\_bytes                            & Total Length of Bwd Packet  \dag          \\ 
\midrule
bidirectional\_duration\_ms               & Flow Duration                         \\ 
\midrule
bidirectional\_min\_ps                    & Packet Length Min  \dag                   \\ 
bidirectional\_max\_ps                    & Packet Length Max  \dag                   \\ 
bidirectional\_mean\_ps                   & Packet Length Mean                    \\ 
bidirectional\_stddev\_ps                 & Packet Length Std                     \\ 
src2dst\_max\_ps                          & Fwd Packet Length Max                 \\ 
src2dst\_min\_ps                          & Fwd Packet Length Min                 \\ 
src2dst\_mean\_ps                         & Fwd Packet Length Mean                \\ 
src2dst\_stddev\_ps                       & Fwd Packet Length Std                 \\ 
dst2src\_max\_ps                          & Bwd Packet Length Max                 \\ 
dst2src\_min\_ps                          & Bwd Packet Length Min                 \\ 
dst2src\_mean\_ps                         & Bwd Packet Length Mean                \\ 
dst2src\_stddev\_ps                       & Bwd Packet Length Std                 \\ 
\midrule
bidirectional\_mean\_piat\_ms             & Flow IAT Mean                         \\ 
bidirectional\_stddev\_piat\_ms           & Flow IAT Std                          \\ 
bidirectional\_max\_piat\_ms              & Flow IAT Max                          \\ 
bidirectional\_min\_piat\_ms              & Flow IAT Min                          \\ 
src2dst\_mean\_piat\_ms                   & Fwd IAT Mean                          \\ 
src2dst\_stddev\_piat\_ms                 & Fwd IAT Std                           \\ 
src2dst\_max\_piat\_ms                    & Fwd IAT Max                           \\ 
src2dst\_min\_piat\_ms                    & Fwd IAT Min                           \\ 
dst2src\_mean\_piat\_ms                   & Bwd IAT Mean                          \\ 
dst2src\_stddev\_piat\_ms                 & Bwd IAT Std                           \\ 
dst2src\_max\_piat\_ms                    & Bwd IAT Max                           \\ 
dst2src\_min\_piat\_ms                    & Bwd IAT Min                           \\ 
\midrule
bidirectional\_fin\_packets               & FIN Flag Count                        \\ 
bidirectional\_syn\_packets               & SYN Flag Count                        \\ 
bidirectional\_rst\_packets               & RST Flag Count                        \\ 
bidirectional\_psh\_packets               & PSH Flag Count                        \\ 
bidirectional\_ack\_packets               & ACK Flag Count                        \\ 
bidirectional\_urg\_packets               & URG Flag Count                        \\ 
bidirectional\_cwr\_packets               & CWR Flag Count  \dag                      \\ 
bidirectional\_ece\_packets               & ECE Flag Count                        \\ 
\midrule
src2dst\_psh\_packets                     & Fwd PSH Flags                         \\ 
dst2src\_psh\_packets                     & Bwd PSH Flags                         \\ 
src2dst\_urg\_packets                     & Fwd URG Flags                         \\ 
dst2src\_urg\_packets                     & Bwd URG Flags                         \\ 
\end{tabular*}
    \begin{tablenotes}
        \item[\dag] There is an inconsistency in the naming conventions used across the original CICIDS-2017 dataset and its subsequent refinements, WTMC-2021 and CRiSIS-2022.
    \end{tablenotes}
    \end{threeparttable}
\end{table}

While feature selection methods could have been employed to reduce the dimensionality of data, we opted to utilize the full feature set provided by the CICIDS-2017 dataset, with specific exclusions aimed at minimizing potential biases. Specifically, the source and destination IP addresses, as well as port numbers flow key attributes were excluded from our feature set. The rationale behind this decision was to avoid biasing the model with information that could inadvertently skew its learning process. For instance, IP addresses present a challenge in terms of transforming them into a format that is effectively utilizable by ML models. On the other hand, including destination port numbers in the feature set could lead the model to associate certain port numbers (e.g., port 21 for FTP or port 22 for SSH) with specific types of attacks (e.g., FTP-Patator or SSH-Patator). This reliance on port numbers could detract from the model's ability to learn and identify attack patterns based on the intrinsic characteristics of the network flows. 

The protocol identifier attribute, however, was retained as part of our feature set. In the CICIDS-2017 dataset, all attacks were executed over the TCP protocol, rendering the inclusion of the protocol identifier somewhat less impactful. Nevertheless, considering that TCP flags were also included in our feature set, we found no compelling reason to exclude the protocol identifier. Its inclusion might still offer marginal benefits in understanding flow characteristics, especially when combined with other features.


It is worth to note that, even for features that appear similar, discrepancies in their calculated values may occur, as also evident in \Cref{tbl:datasets,tbl:anomalies,tbl:fin-rst}. These differences can arise due to variations in how the features are computed and the specific mechanisms employed for updating and terminating flows. Understanding the reasons behind these discrepancies 
is not within the scope of our current work. A more thorough investigation would be required to fully comprehend these variations and their root causes. 

\subsection{Metrics}

In evaluating the performance of the RF algorithm on various dataset versions, we used a comprehensive set of metrics for both binary and multi-class classification scenarios. For binary classification, we consolidated all attack types under a singular ``ANOMALY'' label, whereas, for multi-class classification, we retained the original labels for individual attack types.

The key metrics employed include \textit{precision} (the ratio of correctly predicted positive observations to the total predicted positive observations), \textit{recall} (the ratio of correctly predicted positive observations to all actual positive observations, also known as true positive rate or sensitivity), \textit{accuracy} (the ratio of correctly predicted observations to total observations), \textit{F1 score} (the weighted average of precision and recall), and \textit{Receiver Operating Characteristic (ROC) - Area Under Curve (AUC)} (measuring the model's ability to distinguish between classes). In addition to the metrics mentioned above, we also calculate \textit{normalized confusion matrices} for each model. 



\subsection{Binary Classification Results}

\Cref{tbl:binary-metrics} presents a comparative analysis of the performance metrics for each dataset in our binary classification setup. The table shows the values for precision (Prec.), recall (Rec.), accuracy (Acc.), F1 score (F1), and Area Under the Receiver Operating Characteristic Curve (AUC). Key observations from \Cref{tbl:binary-metrics} are as follows:

\definecolor{cerise}{rgb}{0.87, 0.19, 0.39}
\definecolor{lightblue}{rgb}{0.68, 0.85, 0.9}
\definecolor{ceil}{rgb}{0.57, 0.63, 0.81}
\definecolor{lightergreen}{rgb}{0.8, 1, 0.8}
\definecolor{magicmint}{rgb}{0.67, 0.94, 0.82}
\definecolor{grannysmithapple}{rgb}{0.66, 0.89, 0.63}
\definecolor{celadon}{rgb}{0.67, 0.88, 0.69}

\begin{table}
    \footnotesize
    \centering
    \caption{Performance Metrics Comparison Across Datasets and Days for Binary Classification}
    \label{tbl:binary-metrics}
    \begin{tblr}{
         colspec = {l l *{5}{c}},
         column{1-7} = {colsep=4.2pt}, 
         vline{1,2,3,8}, hline{2,3,6,11,16,21},
         row{1-2} = {font=\bfseries, halign=c},
         rows = {halign=r},
         process=\funcColor{0.7723}{red6}{0.8862}{blue6}{1.0}{lightergreen}{3}{3}
        }
        \toprule
        \SetCell[r=2,c=1]{c} \rotatebox[origin=c]{90}{Day} &
        \SetCell[r=2,c=1]{c} Dataset &
        \SetCell[r=1,c=5]{c} Binary Classification & & & & \\
        & & Prec. & Rec. & Acc. & F1 & AUC \\
        \midrule
        \SetCell[r=5]{c} \rotatebox[origin=c]{90}{\textbf{Tue}}
        & CICIDS-2017 & 0.9984 & 0.7723 & 0.9929 & 0.8709 & 0.8861 \\
        & WTMC-2021 & 1.0000 & 1.0000 & 1.0000 & 1.0000 & 1.0000 \\
        & CRiSIS-2022 & 1.0000 & 0.9986 & 1.0000 & 0.9993 & 0.9993 \\
        & NFS-2023-nTE & 1.0000 & 0.9986 & 1.0000 & 0.9993 & 0.9993 \\
        & NFS-2023-TE & 0.9991 & 0.9997 & 1.0000 & 0.9994 & 0.9998 \\
        \midrule
        \SetCell[r=5]{c} \rotatebox[origin=c]{90}{\textbf{Wed}}
        & CICIDS-2017 & 0.9792 & 0.9958 & 0.9908 & 0.9874 & 0.9918 \\
        & WTMC-2021 & 0.9999 & 0.9998 & 0.9999 & 0.9999 & 0.9999 \\
        & CRiSIS-2022 & 1.0000 & 0.9999 & 1.0000 & 1.0000 & 1.0000 \\
        & NFS-2023-nTE & 1.0000 & 0.9997 & 0.9999 & 0.9998 & 0.9999 \\
        & NFS-2023-TE & 0.9998 & 0.9995 & 0.9998 & 0.9996 & 0.9997 \\
        \midrule
        \SetCell[r=5]{c} \rotatebox[origin=c]{90}{\textbf{Thu}}
        & CICIDS-2017 & 0.9966 & 0.8902 & 0.9995 & 0.9404 & 0.9451 \\ 
        & WTMC-2021 & 1.0000 & 0.8806 & 0.9999 & 0.9365 & 0.9403 \\
        & CRiSIS-2022 & 0.9685 & 0.9987 & 0.9965 & 0.9834 & 0.9975 \\
        & NFS-2023-nTE & 0.9834 & 0.9995 & 0.9969 & 0.9914 & 0.9979 \\
        & NFS-2023-TE & 0.9840 & 0.9027 & 0.9835 & 0.9416 & 0.9501 \\
        \midrule
        \SetCell[r=5]{c} \rotatebox[origin=c]{90}{\textbf{Fri}}
        & CICIDS-2017 & 0.9992 & 0.9962 & 0.9981 & 0.9977 & 0.9978 \\
        & WTMC-2021 & 0.9976 & 0.9997 & 0.9987 & 0.9986 & 0.9988 \\
        & CRiSIS-2022 & 0.9978 & 0.9997 & 0.9988 & 0.9987 & 0.9989 \\
        & NFS-2023-nTE & 1.0000 & 0.9999 & 1.0000 & 1.0000 & 1.0000 \\
        & NFS-2023-TE & 1.0000 & 0.9993 & 0.9998 & 0.9996 & 0.9996 \\
        \bottomrule
    \end{tblr}
\end{table}

\begin{table*}
    \newcommand{\myrotation}{90}
    \centering
    \caption{Day-wise and Dataset-wise Confusion Matrix Comparison for Binary Classification}
    \label{tbl:binary-confusion_matrices}
    \begin{tabular*}{\tblwidth}{@{}CCCCC@{}}
          & Tuesday & Wednesday & Thursday & Friday \\
          & & & & \\
        \rotatebox[origin=l]{\myrotation}{~~~~~CICIDS-2017}  
        & \includegraphics[width=0.2\linewidth]{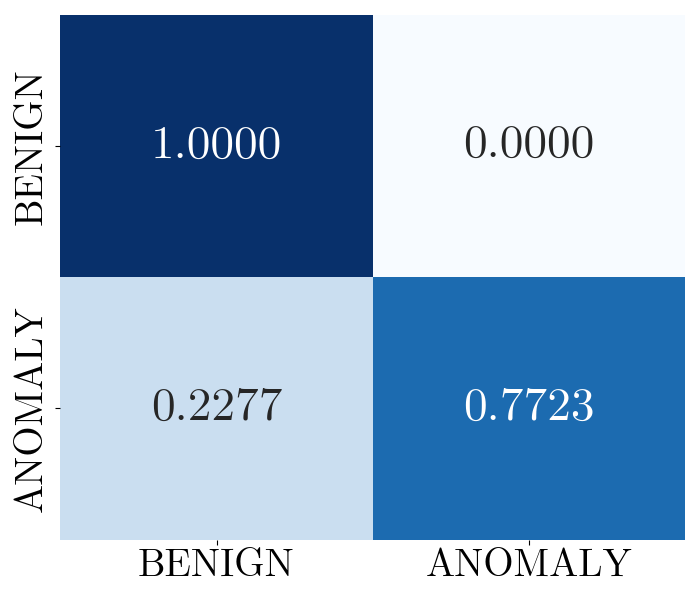} 
        & \includegraphics[width=0.2\linewidth]{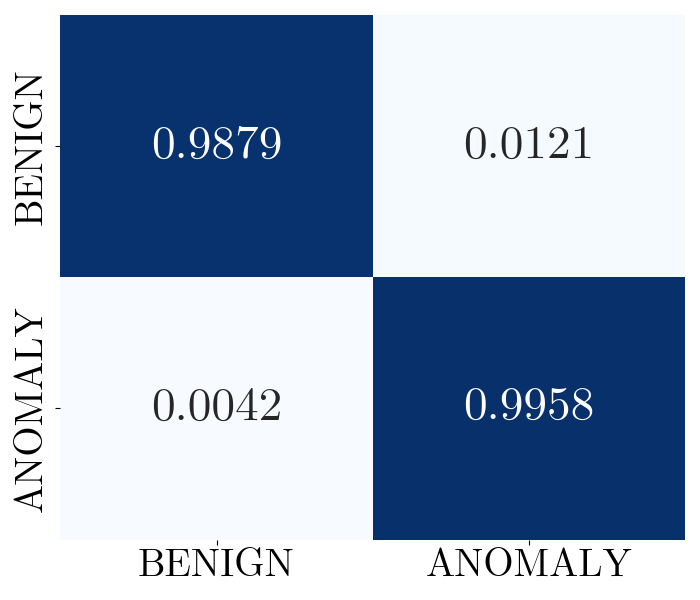} 
        & \includegraphics[width=0.2\linewidth]{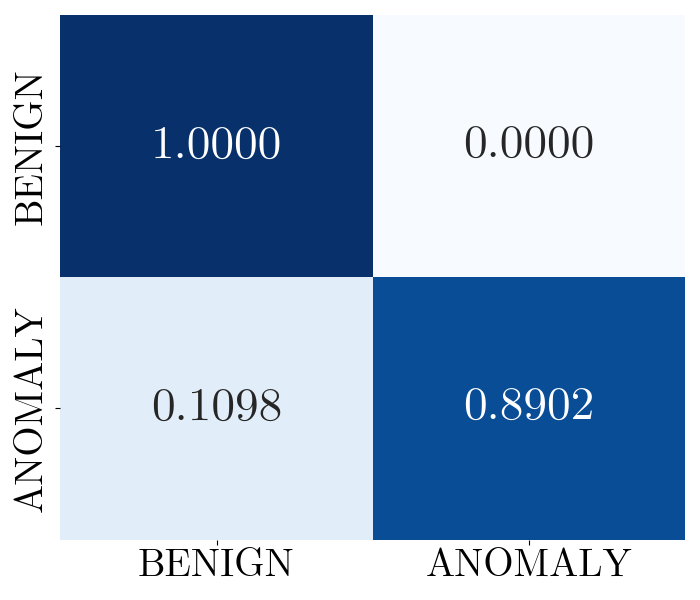} 
        & \includegraphics[width=0.2\linewidth]{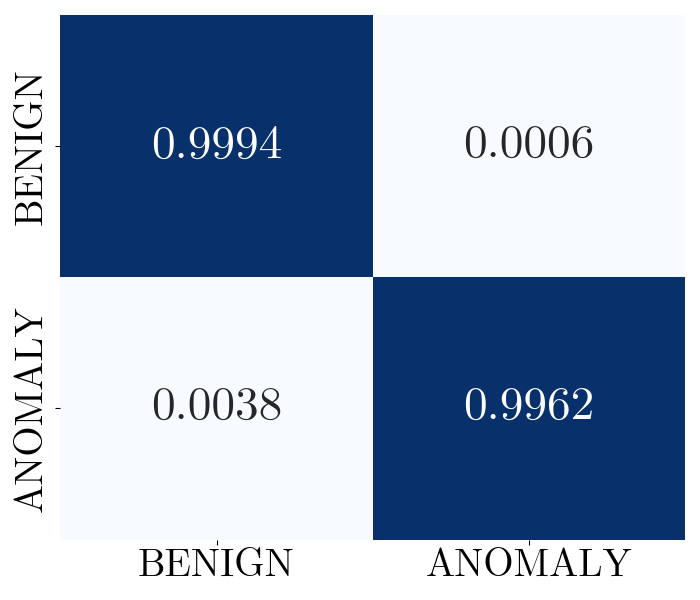}
        \\
        \rotatebox[origin=l]{\myrotation}{~~~~~WTMC-2021}  
        & \includegraphics[width=0.2\linewidth]{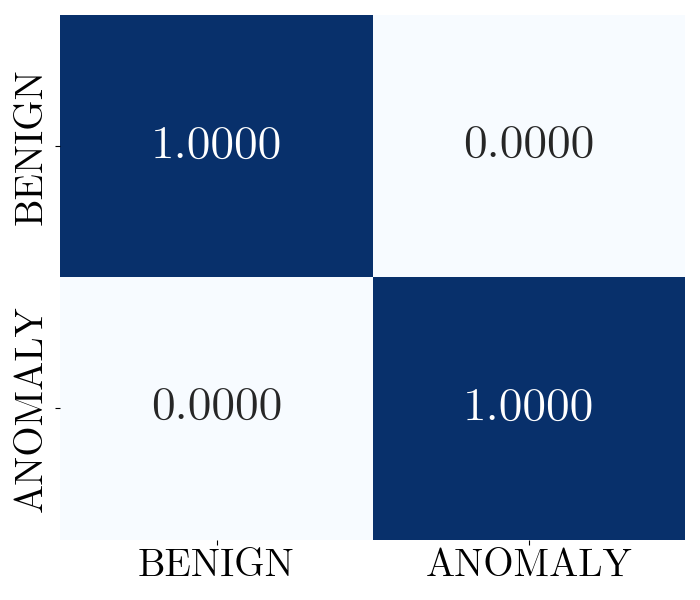} 
        & \includegraphics[width=0.2\linewidth]{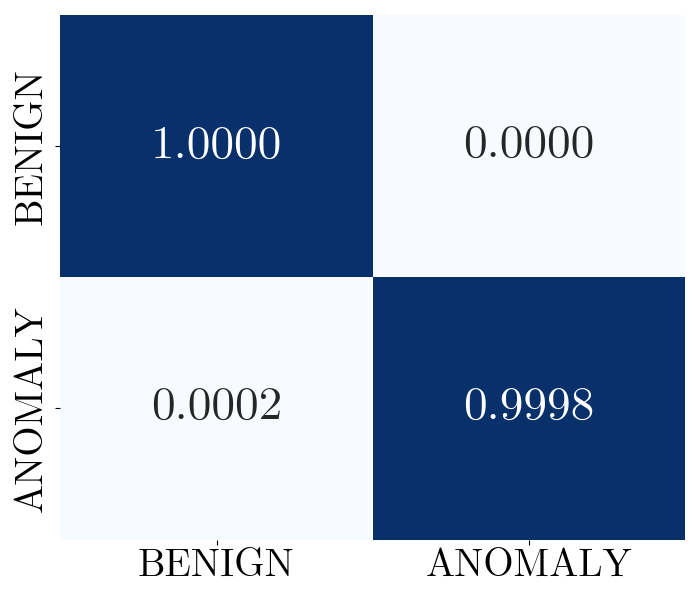} 
        & \includegraphics[width=0.2\linewidth]{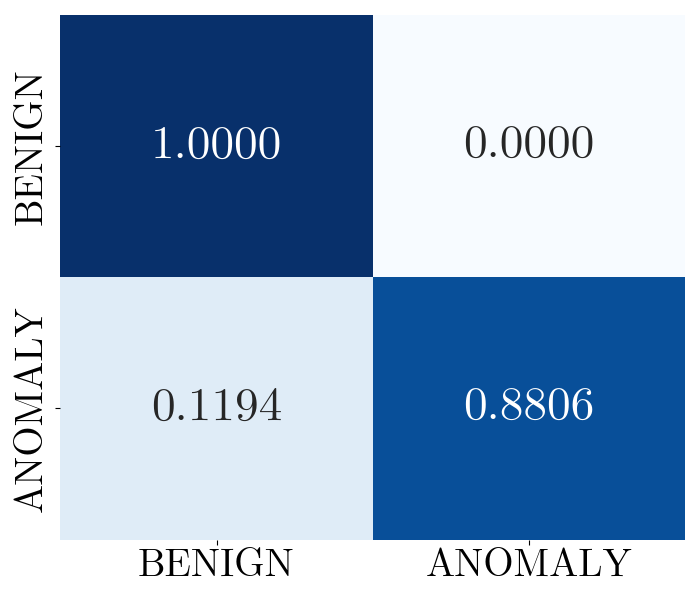} 
        & \includegraphics[width=0.2\linewidth]{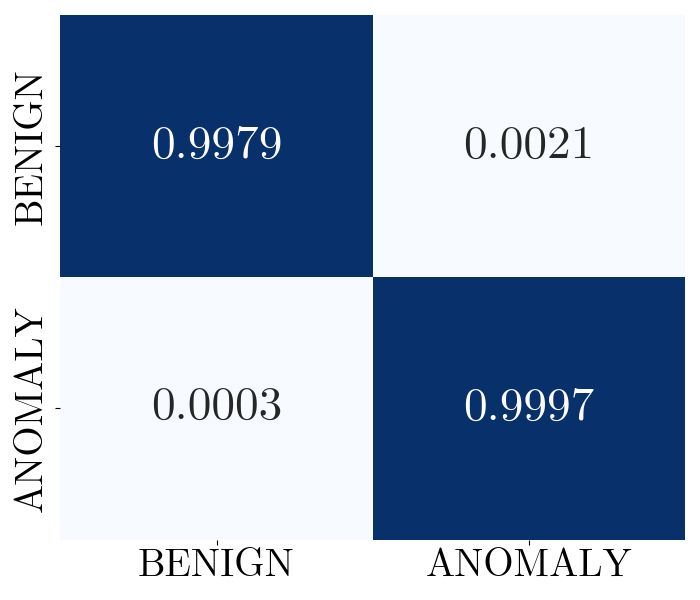}
        \\
        \rotatebox[origin=l]{\myrotation}{~~~~~CRiSIS-2022}  
        & \includegraphics[width=0.2\linewidth]{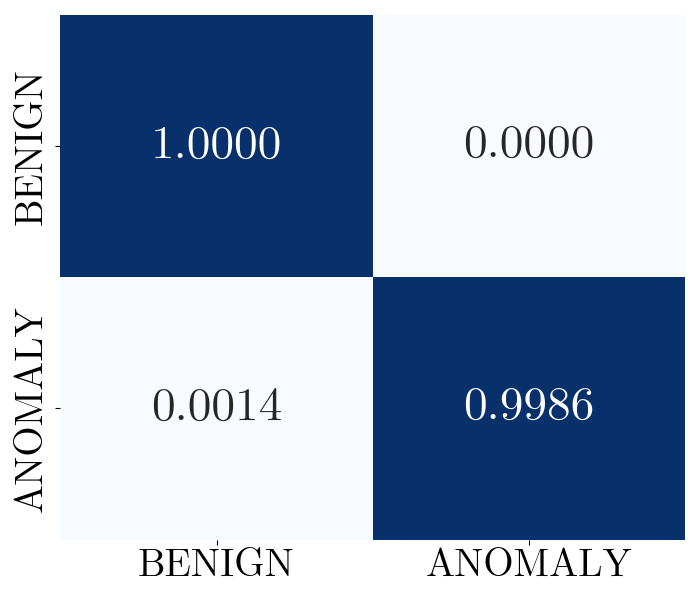} 
        & \includegraphics[width=0.2\linewidth]{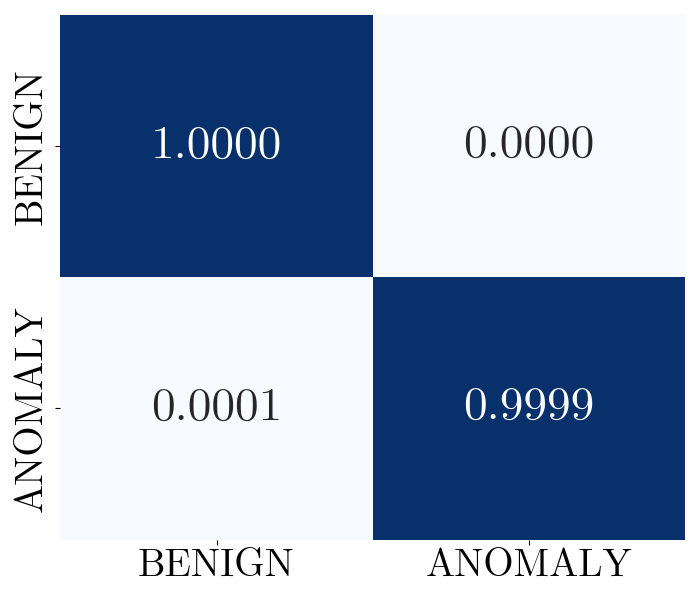} 
        & \includegraphics[width=0.2\linewidth]{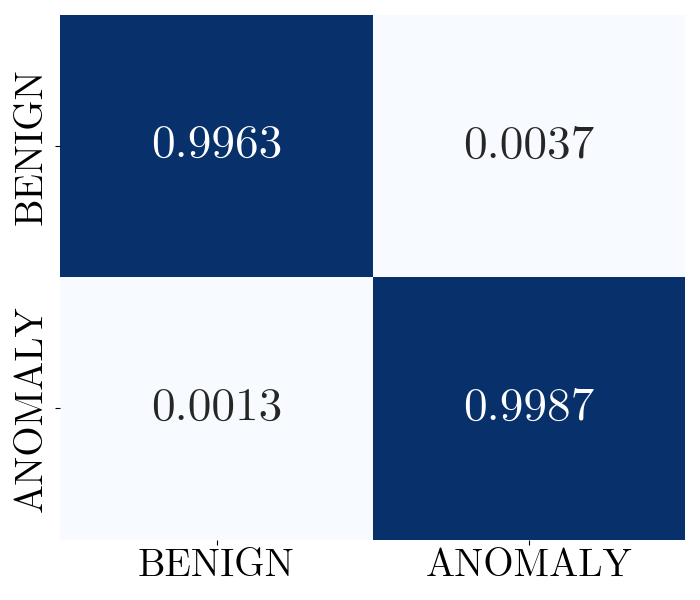} 
        & \includegraphics[width=0.2\linewidth]{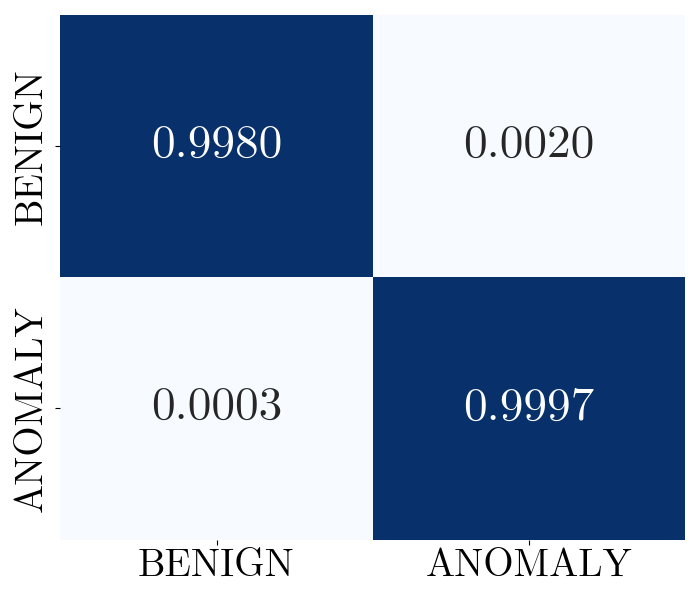}
        \\
        \rotatebox[origin=l]{\myrotation}{~~~~NFS-2023-nTE}  
        & \includegraphics[width=0.2\linewidth]{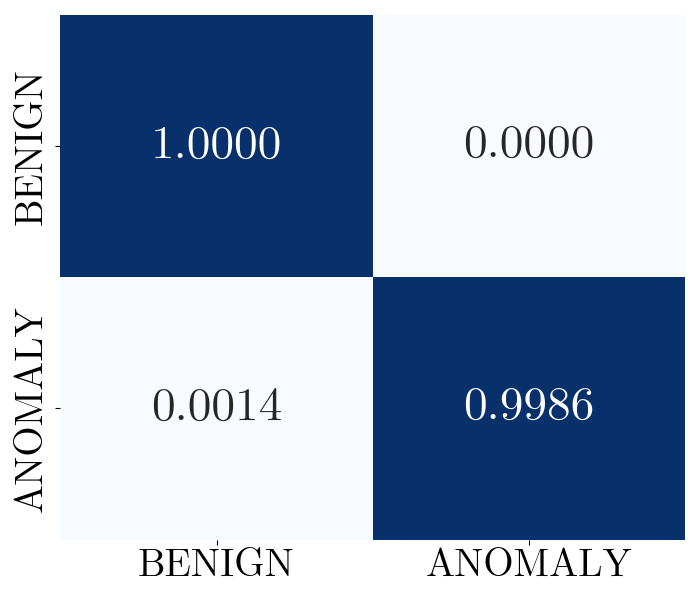} 
        & \includegraphics[width=0.2\linewidth]{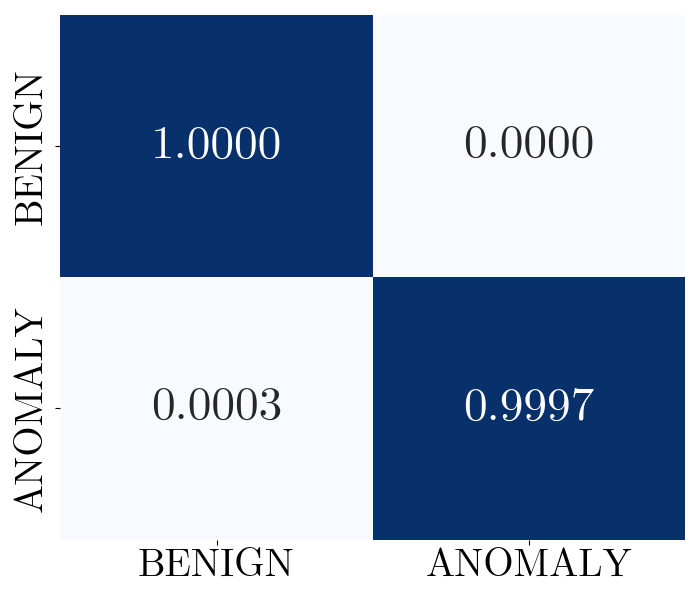} 
        & \includegraphics[width=0.2\linewidth]{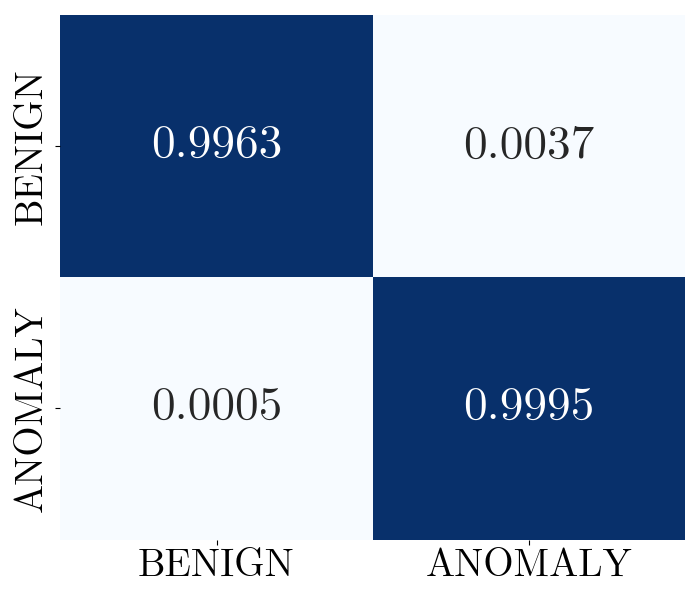} 
        & \includegraphics[width=0.2\linewidth]{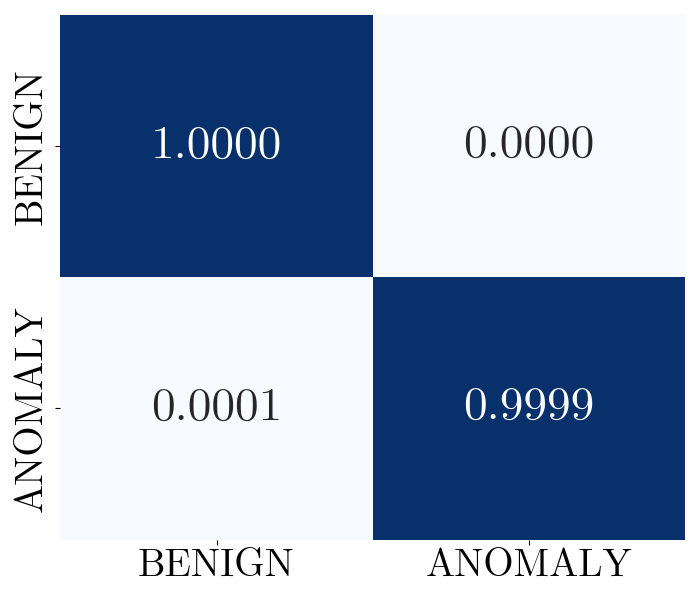}
        \\
        \rotatebox[origin=l]{\myrotation}{~~~~~NFS-2023-TE}  
        & \includegraphics[width=0.2\linewidth]{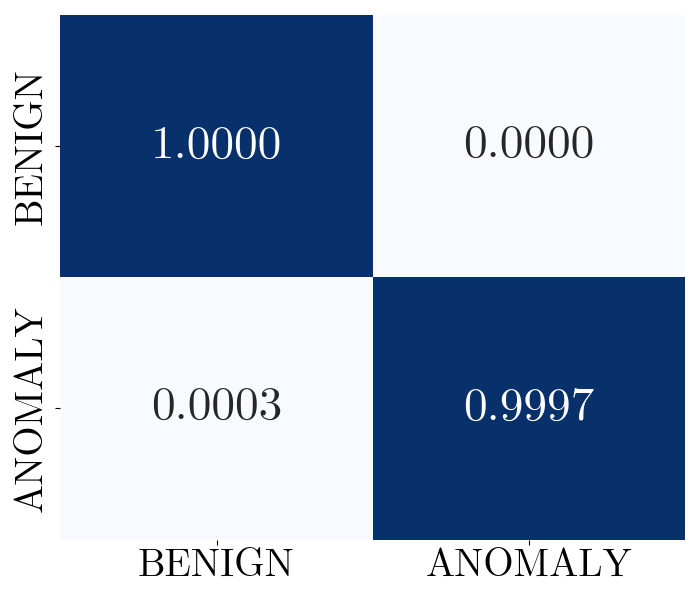} 
        & \includegraphics[width=0.2\linewidth]{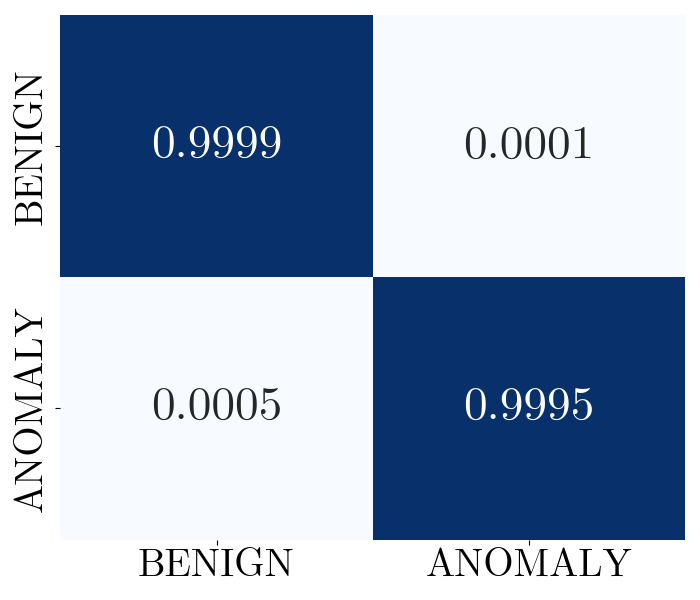} 
        & \includegraphics[width=0.2\linewidth]{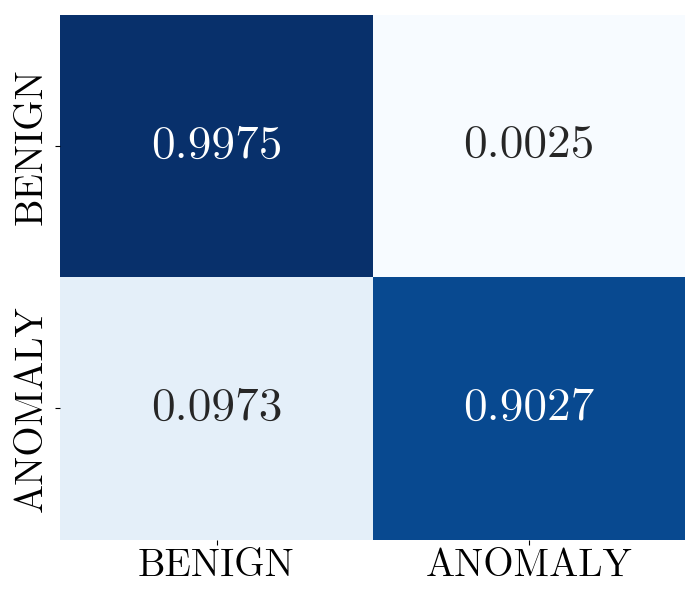} 
        & \includegraphics[width=0.2\linewidth]{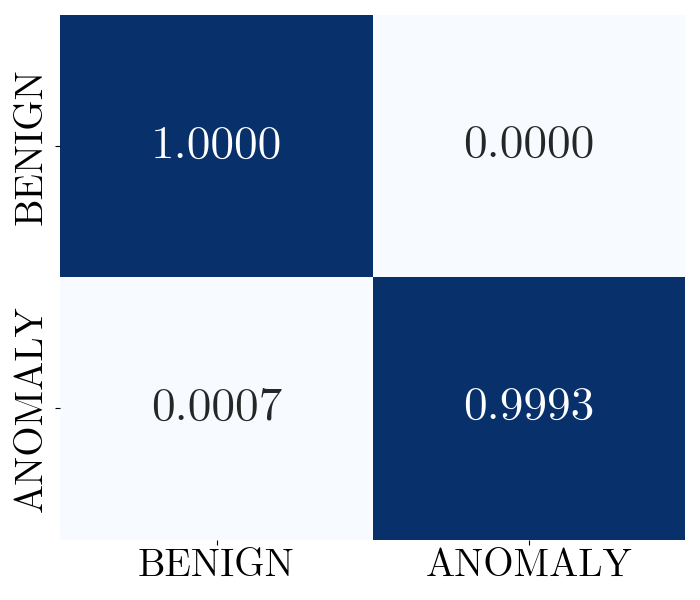}
        \\
    \end{tabular*}
\end{table*}

\begin{itemize}
    \item 
    Both WTMC-2021 and CRiSIS-2022 datasets consistently show near-perfect scores across all metrics. This indicates exceptional model performance in correctly classifying flows, with minimal misclassifications.
    \item 
    The CICIDS-2017 dataset exhibits variability in performance across different days, with particularly lower recall and F1 scores on Tuesday and Thursday. 
    This suggests discrepancies in data quality or representativeness, impacting the model's ability to detect true positives effectively.
    \item 
    An interesting pattern emerges with the Thursday datasets. While accuracy is consistently high across all versions, there are noticeable variations in the other metrics. CICIDS-2017, WTMC-2021, and NFS-2023-TE demonstrate lower recall, F1 scores, and ROC AUC, indicating some challenges in consistently identifying true positives. In contrast, CRiSIS-2022 and NFS-2023-nTE achieve higher metrics, reflecting a more balanced detection capability. This suggests that the Thursday dataset may have unique characteristics influencing the models' predictive performance.
    \item 
    NFS-2023-nTE and NFS-2023-TE datasets generally align closely with the refined datasets (WTMC-2021 and CRiSIS-2022) in terms of performance metrics. This indicates that the methodological improvements in CICFlowMeter are effective.
    \item 
    The NFS-2023-TE dataset, particularly on Friday, shows high scores in all metrics, aligning closely with the NFS-2023-nTE dataset. This suggests that the TCP flag-based expiration policy in NFS-2023-TE does not adversely affect the model's ability to classify anomalies accurately.
    \item 
    The NFS-2023-nTE dataset demonstrates excellent precision and recall, particularly on Wednesday and Friday, indicating its robustness in identifying true positives without increasing false positives.
\end{itemize}


Complementary to the performance metrics detailed in \Cref{tbl:binary-metrics}, the confusion matrices in \Cref{tbl:binary-confusion_matrices} provide additional insights into the classification patterns of each model. The key findings that can be observed:

\begin{itemize}
    \item 
    The confusion matrices across all datasets for the days of Tuesday (with an exception for CICIDS-2017), Wednesday, and Friday show that the True Positive Rates (the recall for the anomaly class) are exceedingly high, often reaching perfect or near-perfect scores. This suggests that the models are highly effective at identifying anomalous flows on these days.
    \item 
    Unlike other days, the Thursday matrices reveal a notably lower True Positive Rate for anomaly detection in certain datasets, such as CICIDS-2017, WTMC-2021, and NFS-2023-TE. This indicates that the models struggle to consistently identify anomalies on this day, which may be due to the unique characteristics or distribution of the Thursday data.
    \item 
    On Tuesday and Wednesday, the models using the WTMC-2021, CRiSIS-2022, and NFS-2023-nTE datasets display perfect classification for benign flows, with zero false positives. This demonstrates the models' strong ability to correctly identify normal behavior without mistakenly flagging it as anomalous.
    \item 
    There is a notable consistency in the performance of the CRiSIS-2022 and NFS-2023-nTE datasets across all days, as evidenced by the high scores for both benign and anomaly classes, indicating stable model performance regardless of the day.
    \item 
    The NFS-2023-nTE dataset stands out on Friday with perfect scores, reflecting the model's accuracy in distinguishing between benign and anomalous flows without error.
\end{itemize}

\begin{figure*}[!h]
	\centering
	\includegraphics[width=0.95\linewidth, trim={0 1.5cm 0 2.5cm},clip]{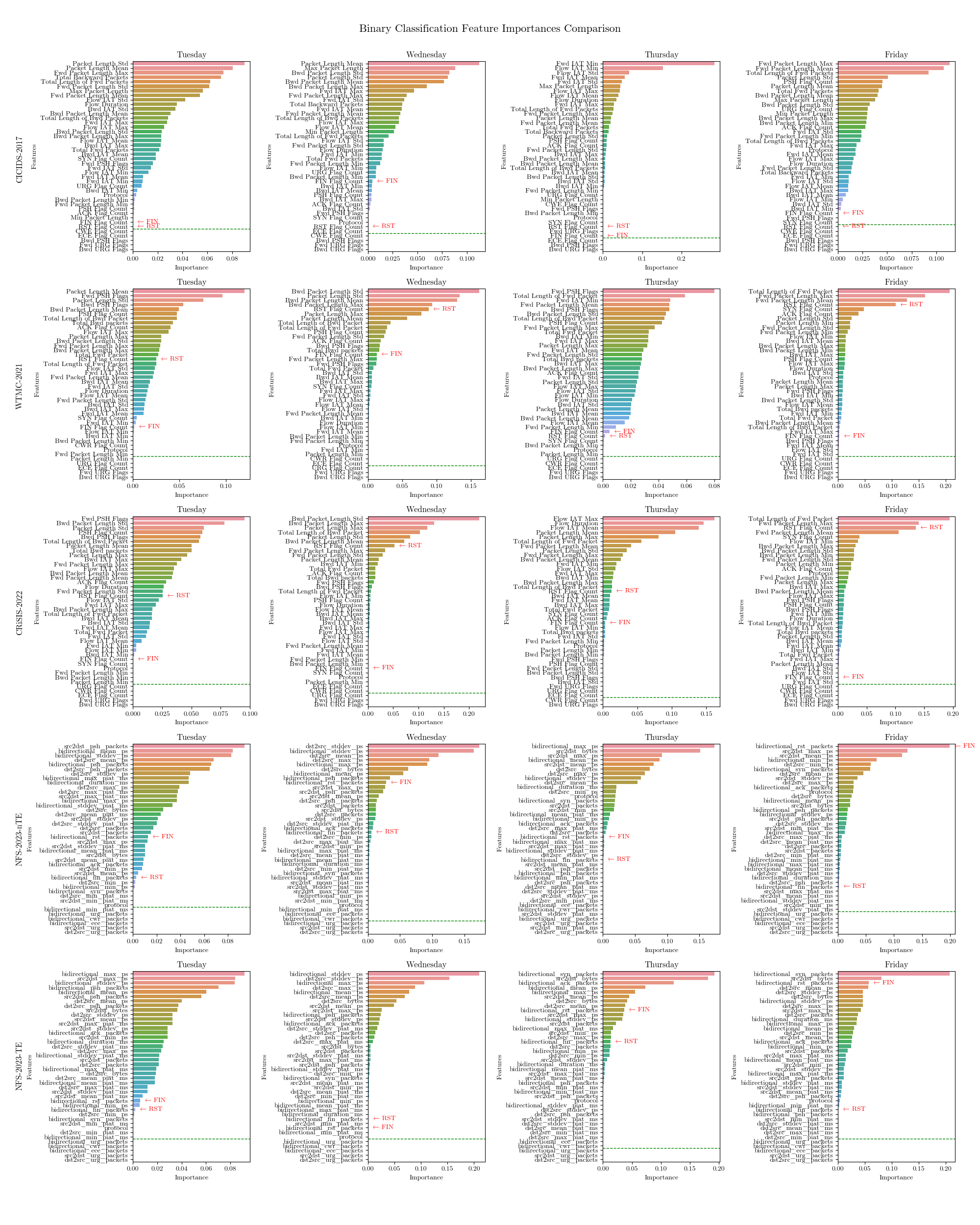}
	  \caption{Binary classification feature importance comparison.}\label{fig:fi-b}
\end{figure*}




The analysis of feature importances, as shown in \Cref{fig:fi-b} for each model, reveals a further insight: TCP FIN and RST flags typically do not emerge as dominant features influencing model outcomes. This trend holds across various dataset versions, with few exceptions. This observation clarifies that the implementation of TCP expiration policies has a minimal impact on the predictive capabilities of the Random Forest model. Instead, the model utilizes a diverse array of features to effectively differentiate between benign and anomalous traffic.

Friday datasets stand out, with TCP FIN and RST flags gaining more prominence. However, even with these features assuming greater importance, the RF classifier maintains high accuracy. This underscores the algorithm's ability to adapt and extract relevant patterns from complex data structures, highlighting its resilience to potential inaccuracies in specific feature measurements.



\subsection{Multi-class Classification Results}

The comparative analysis of multi-class classification performance across different versions of the CICIDS-2017 dataset is depicted in \Cref{tbl:multi-metrics}. Notable observations from the table include:

\begin{itemize}
    \item 
    The WTMC-2021, CRiSIS-2022, NFS-2023-nTE, and NFS-2023-TE datasets exhibit near-perfect or perfect scores across all metrics on Tuesday, Wednesday, and Friday, indicating an exceptional ability of the models to classify multiple attack types with high precision and reliability.
    \item 
    The original CICIDS-2017 dataset shows slightly lower performance compared to the other datasets, particularly on Tuesday and Wednesday, which could suggest some limitations in the dataset's consistency or potential issues with class imbalances affecting model performance.
    \item 
    Both NFS-2023-nTE and NFS-2023-TE datasets achieve perfect scores in all metrics on Tuesday and Friday, and near-perfect scores on Wednesday and Thursday. This indicates that the datasets generated using NFStream, with and without TCP flag-based flow expiration, provide robust bases for model training and evaluation.
    \item 
    On Thursday, there is a noticeable performance drop for the CICIDS-2017 and NFS-2023-TE datasets compared to other days and datasets. This suggests some difficulty in distinguishing between classes on this particular day.
    \item 
    Across all datasets and days, the models maintain high accuracy levels, demonstrating their effectiveness in correctly classifying both benign and anomalous flows in a multi-class setting.
\end{itemize}

\begin{table}
    \footnotesize
    \caption{Performance Metrics Comparison Across Datasets and Days for Multi-class Classification}
    \label{tbl:multi-metrics}
    \begin{tblr}{
         colspec = {l l *{5}{c}},
         column{1-7} = {colsep=4.2pt}, 
         vline{1,2,3,8}, hline{2,3,6,11,16,21},
         row{1-2} = {font=\bfseries, halign=c},
         rows = {halign=r},
         process=\funcColor{0.7723}{red6}{0.8862}{blue6}{1.0}{lightergreen}{3}{3}
    }
        \toprule
        \SetCell[r=2,c=1]{c} \rotatebox[origin=c]{90}{Day} &
        \SetCell[r=2,c=1]{c} Dataset &
        \SetCell[r=1,c=5]{c} Multi-class Classification & & & &  \\ 
        & & Prec. & Rec. & Acc. & F1 & AUC \\
        \midrule
        \SetCell[r=5]{c} \rotatebox[origin=c]{90}{\textbf{Tue}}
        & CICIDS-2017 & 0.9935 & 0.9935 & 0.9935 & 0.9925 & 0.9949 \\
        & WTMC-2021 & 0.9999 & 0.9999 & 0.9999 & 0.9999 & 1.0000 \\
        & CRiSIS-2022 & 0.9999 & 0.9999 & 0.9999 & 0.9999 & 0.9998 \\
        & NFS-2023-nTE & 1.0000 & 1.0000 & 1.0000 & 1.0000 & 1.0000 \\
        & NFS-2023-TE & 1.0000 & 1.0000 & 1.0000 & 1.0000 & 1.0000 \\
        \midrule
        \SetCell[r=5]{c} \rotatebox[origin=c]{90}{\textbf{Wed}}
        & CICIDS-2017 & 0.9907 & 0.9906 & 0.9906 & 0.9906 & 0.9989 \\
        & WTMC-2021 & 0.9999 & 0.9999 & 0.9999 & 0.9999 & 1.0000 \\
        & CRiSIS-2022 & 0.9999 & 0.9999 & 0.9999 & 0.9999 & 1.0000 \\
        & NFS-2023-nTE & 0.9997 & 0.9997 & 0.9997 & 0.9997 & 1.0000 \\
        & NFS-2023-TE & 0.9997 & 0.9997 & 0.9997 & 0.9997 & 0.9999 \\
        \midrule
        \SetCell[r=5]{c} \rotatebox[origin=c]{90}{\textbf{Thu}}
        & CICIDS-2017 & 0.9979 & 0.9981 & 0.9981 & 0.9979 & 0.9840 \\
        & WTMC-2021 & 0.9999 & 0.9999 & 0.9999 & 0.9999 & 0.9925 \\
        & CRiSIS-2022 & 0.9966 & 0.9965 & 0.9965 & 0.9965 & 0.9986 \\
        & NFS-2023-nTE & 0.9970 & 0.9969 & 0.9969 & 0.9969 & 0.9990 \\
        & NFS-2023-TE & 0.9839 & 0.9839 & 0.9839 & 0.9836 & 0.9954 \\
        \midrule
        \SetCell[r=5]{c} \rotatebox[origin=c]{90}{\textbf{Fri}}
        & CICIDS-2017 & 0.9977 & 0.9979 & 0.9979 & 0.9977 & 0.9999 \\
        & WTMC-2021 & 0.9987 & 0.9987 & 0.9987 & 0.9987 & 0.9998 \\
        & CRiSIS-2022 & 0.9988 & 0.9988 & 0.9988 & 0.9988 & 0.9998 \\
        & NFS-2023-nTE & 1.0000 & 1.0000 & 1.0000 & 1.0000 & 1.0000 \\
        & NFS-2023-TE & 0.9998 & 0.9998 & 0.9998 & 0.9998 & 1.0000 \\
        \bottomrule
    \end{tblr}
\end{table}

Complementary to the performance metrics in \Cref{tbl:multi-metrics}, the confusion matrices in \Cref{tbl:multi-confusion_matrices} shed light on the classification performance for individual attack types across the datasets on different days. The matrices reveal several noteworthy patterns:

\begin{itemize}
    \item 
    Across all datasets, certain attack types such as FTP-Patator and SSH-Patator on Tuesday and Infiltration on Wednesday are classified with high precision. This indicates that the models are particularly effective at detecting these types of attacks with minimal false positives.
    \item 
    There is some variation in the recall rates for different attack types. For example, on Thursday, the recall for DoS attacks is noticeably lower in the CICIDS-2017 dataset compared to the others. This variation suggests that some datasets or models may be better tuned to recognize specific attacks over others.
    \item 
    The NFS-2023-nTE and NFS-2023-TE datasets demonstrate consistently high precision and recall across all attack types. This uniformity highlights the robustness of the NFStream tool in generating datasets that lead to reliable classification models.
    \item 
    Similar to the findings in \Cref{tbl:binary-metrics,tbl:binary-confusion_matrices,tbl:multi-metrics}, the confusion matrices for Thursday reveal some challenges in classifying attacks, particularly for the NFS-2023-TE dataset. The NFS-2023-TE dataset displays slightly less precision and recall for certain attack types, aligning with the earlier observation of lower performances on this day.
    \item 
    Despite some day-to-day and attack-type-specific variations, the overall accuracy across all datasets remains high. This suggests that the RF models are generally well-calibrated to distinguish between benign traffic and various types of network anomalies.
\end{itemize}

\begin{figure*}[!h]
	\centering
	\includegraphics[width=0.95\linewidth, trim={0 1.5cm 0 2.5cm},clip]{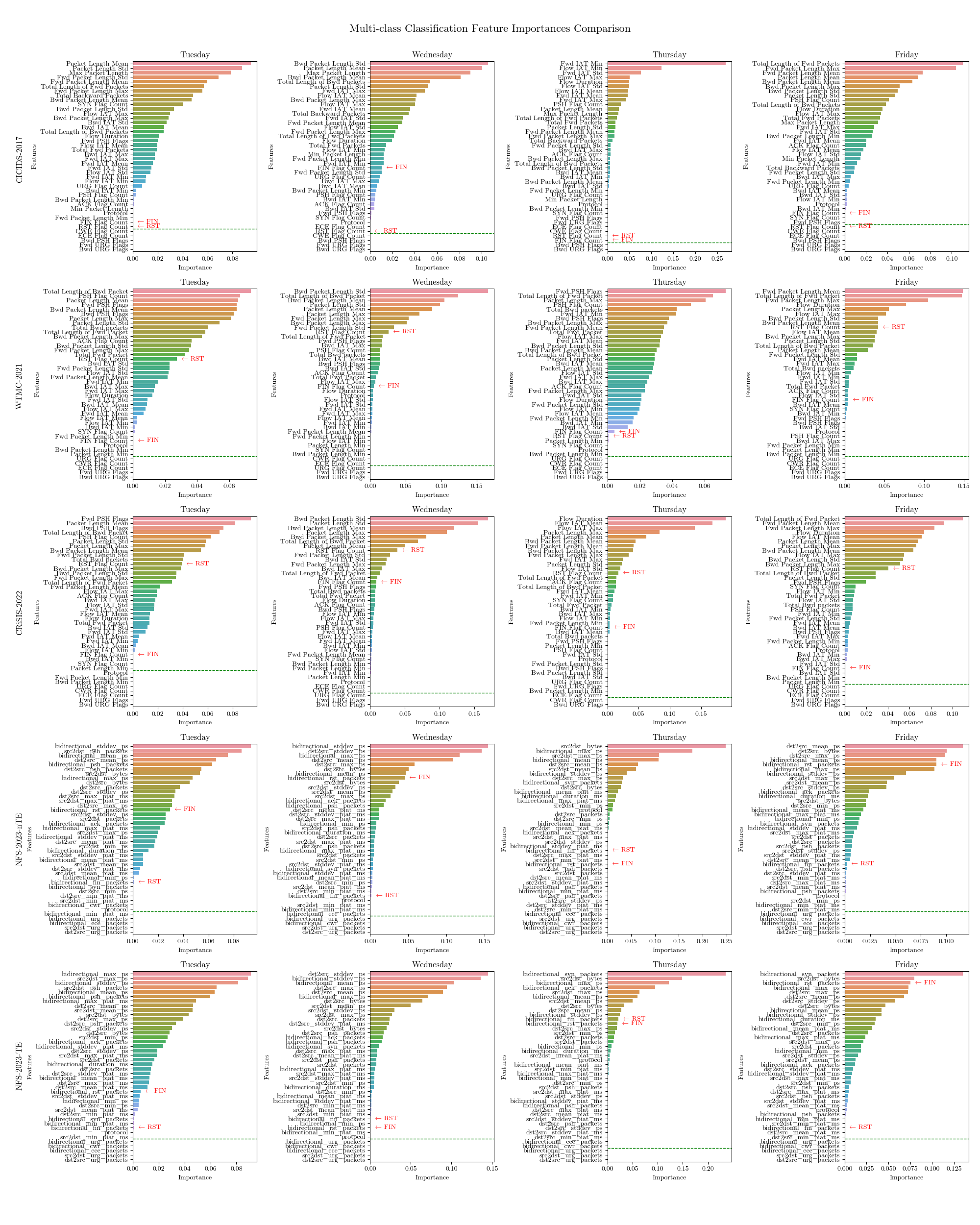}
	  \caption{Multi-class classification feature importance comparison.}\label{fig:fi-m}
\end{figure*}

Feature importances for the multi-class scenarios, as illustrated in \Cref{fig:fi-m}, exhibit trends consistent with those observed in the binary classification scenarios (\Cref{fig:fi-b}). In these analyses, TCP FIN and RST flags generally do not stand out as dominant features affecting model outcomes. Similar to the binary results, these flags become slightly more pronounced in the datasets from Friday, yet even when these features assume greater significance, the RF classifier continues to demonstrate high accuracy. This consistency highlights the classifier's robustness, effectively leveraging a wide range of features to accurately distinguish between multiple classes of network traffic. Such adaptability is crucial, demonstrating the model’s ability to maintain performance reliability despite variations in feature importance across different types of data contexts.

\begin{table*}
    \newcommand{\myrotation}{90}
    \centering
    \caption{Day-wise and Dataset-wise Confusion Matrix Comparison for Multi-class Classification}
    \label{tbl:multi-confusion_matrices}
    \begin{tabular*}{\tblwidth}{@{}CCCCC@{}}
          & Tuesday & Wednesday & Thursday & Friday \\
          & & & & \\
        \rotatebox[origin=l]{\myrotation}{~~~~~CICIDS-2017}  
        & \includegraphics[width=0.2\linewidth]{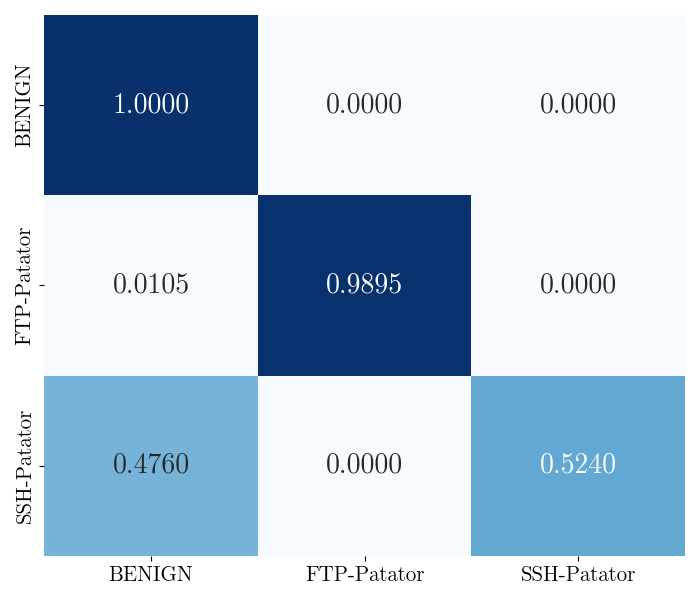} 
        & \includegraphics[width=0.2\linewidth]{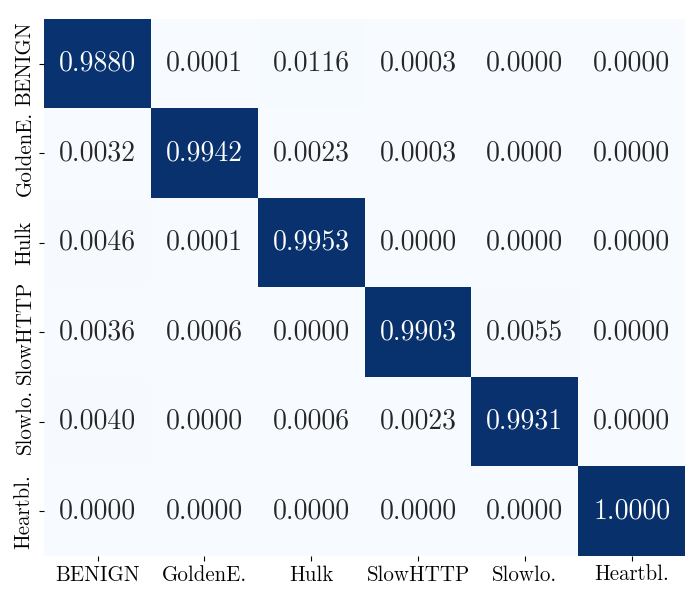} 
        & \includegraphics[width=0.2\linewidth]{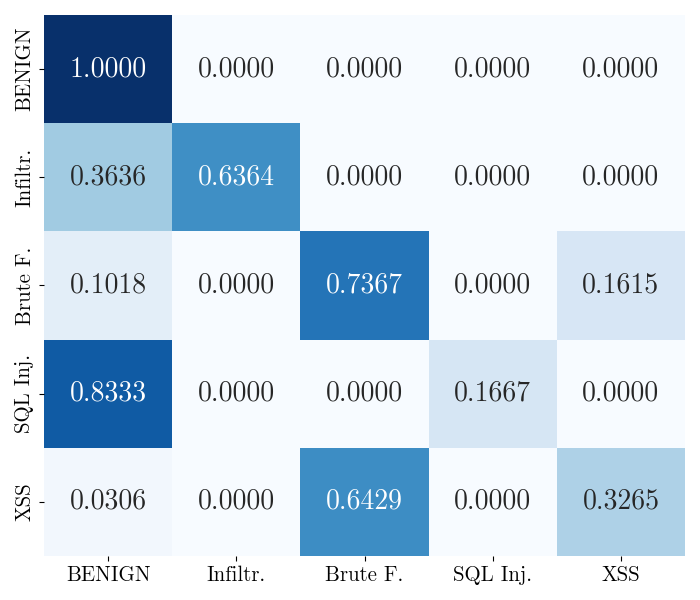} 
        & \includegraphics[width=0.2\linewidth]{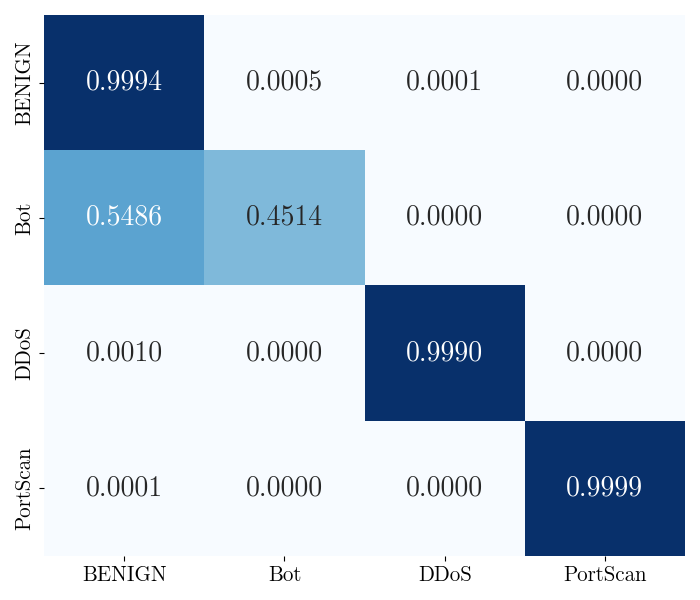}
        \\
        \rotatebox[origin=l]{\myrotation}{~~~~~WTMC-2021}  
        & \includegraphics[width=0.2\linewidth]{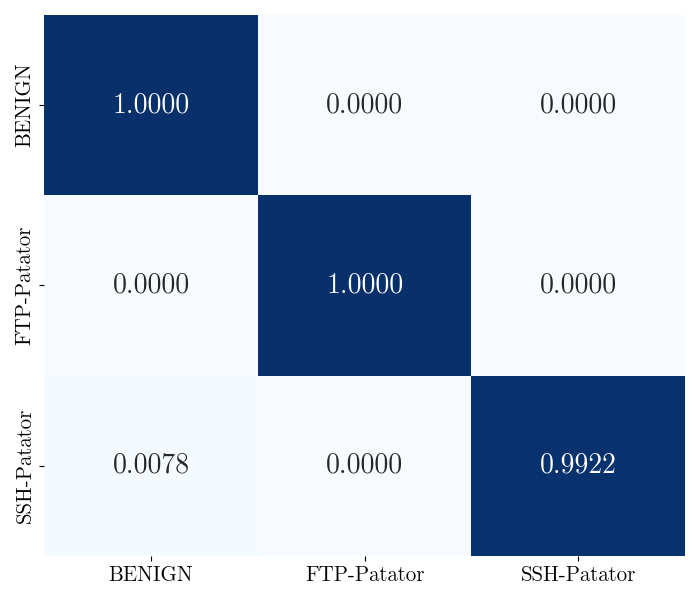} 
        & \includegraphics[width=0.2\linewidth]{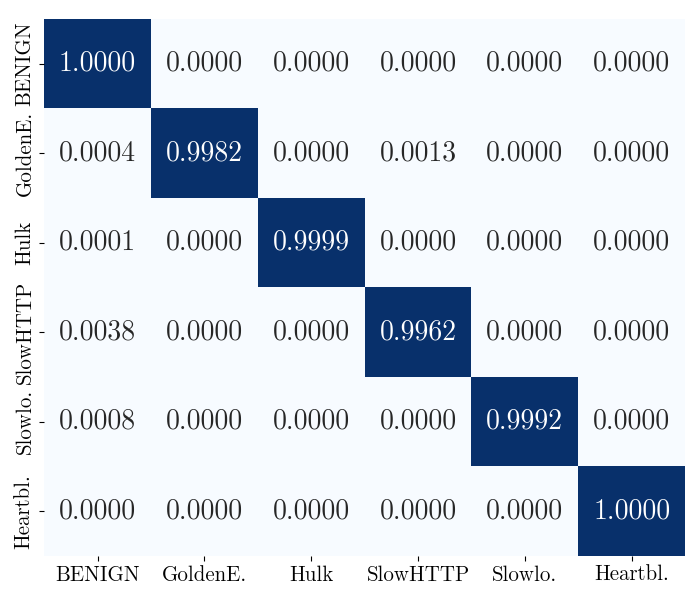} 
        & \includegraphics[width=0.2\linewidth]{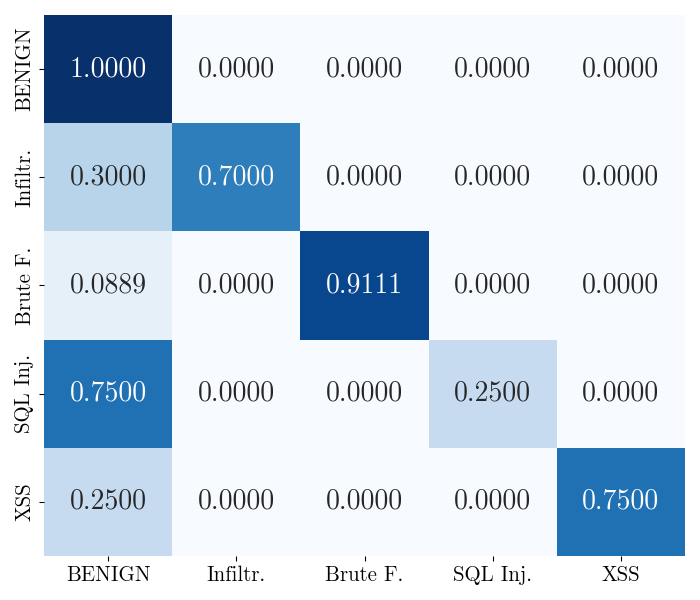} 
        & \includegraphics[width=0.2\linewidth]{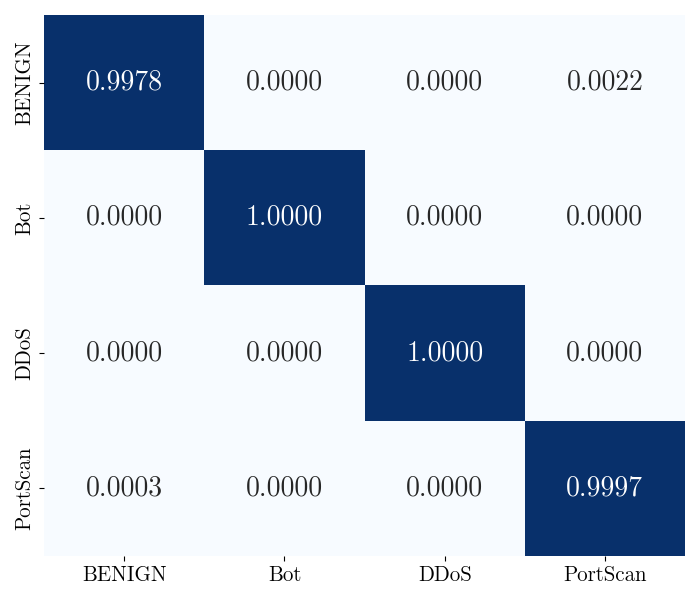}
        \\
        \rotatebox[origin=l]{\myrotation}{~~~~~CRiSIS-2022}  
        & \includegraphics[width=0.2\linewidth]{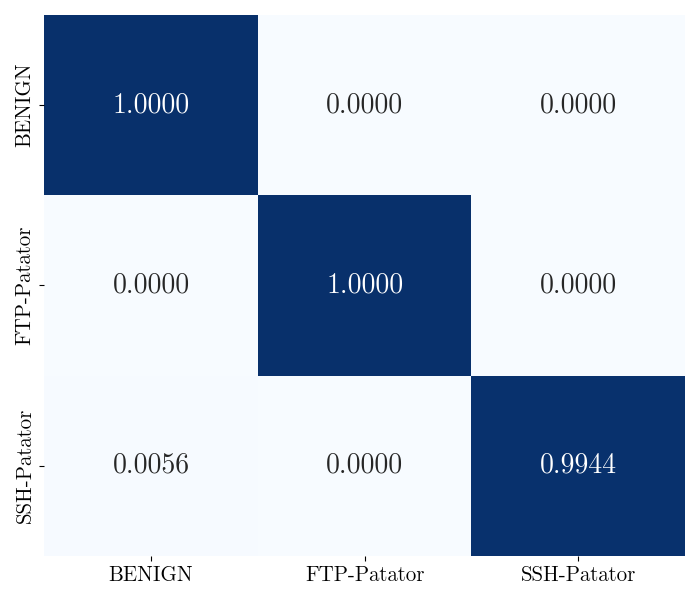} 
        & \includegraphics[width=0.2\linewidth]{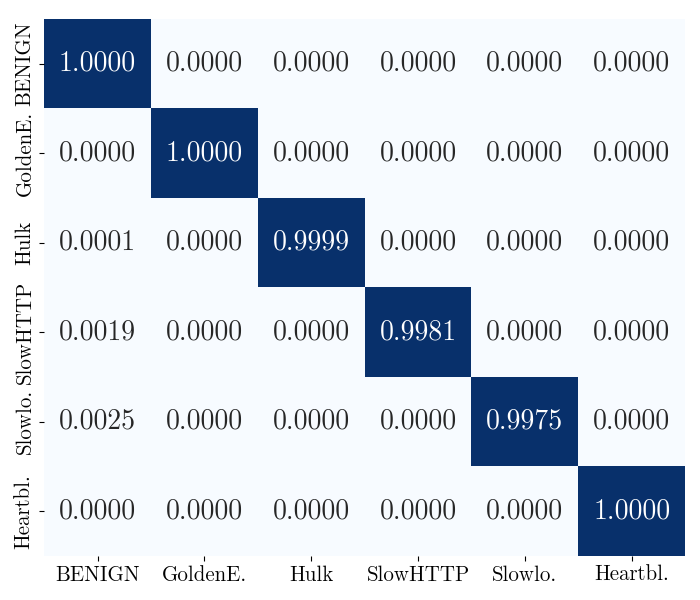} 
        & \includegraphics[width=0.2\linewidth]{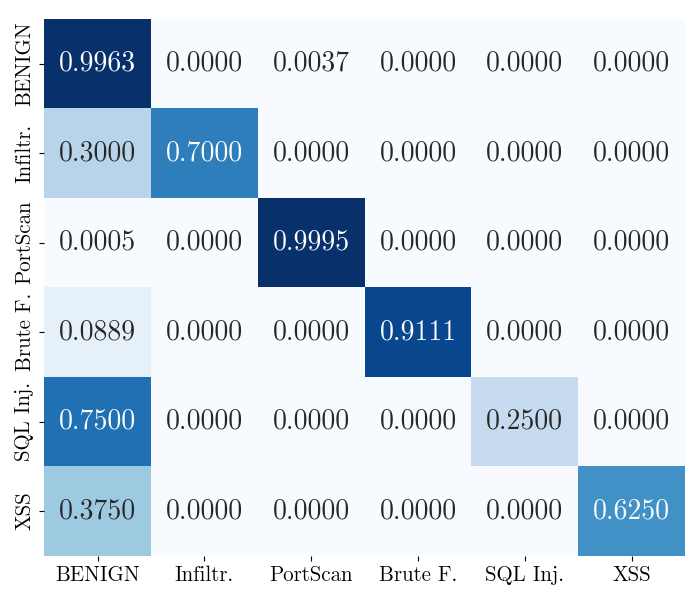} 
        & \includegraphics[width=0.2\linewidth]{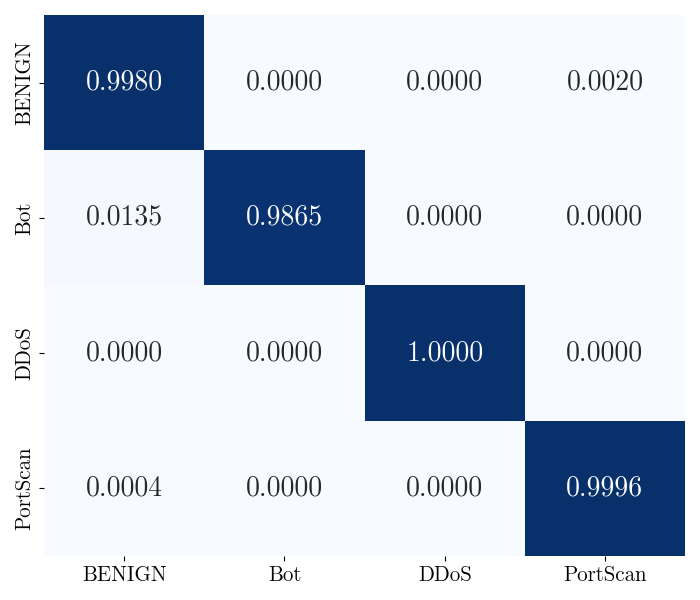}
        \\
        \rotatebox[origin=l]{\myrotation}{~~~~NFS-2023-nTE}  
        & \includegraphics[width=0.2\linewidth]{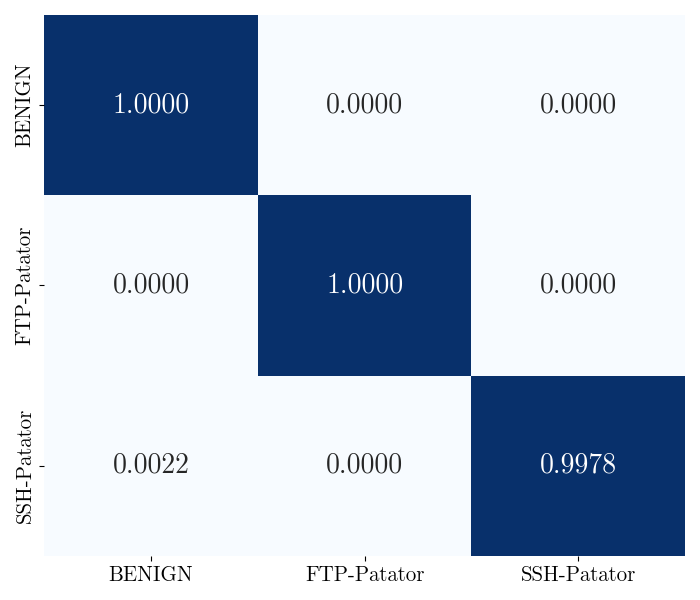} 
        & \includegraphics[width=0.2\linewidth]{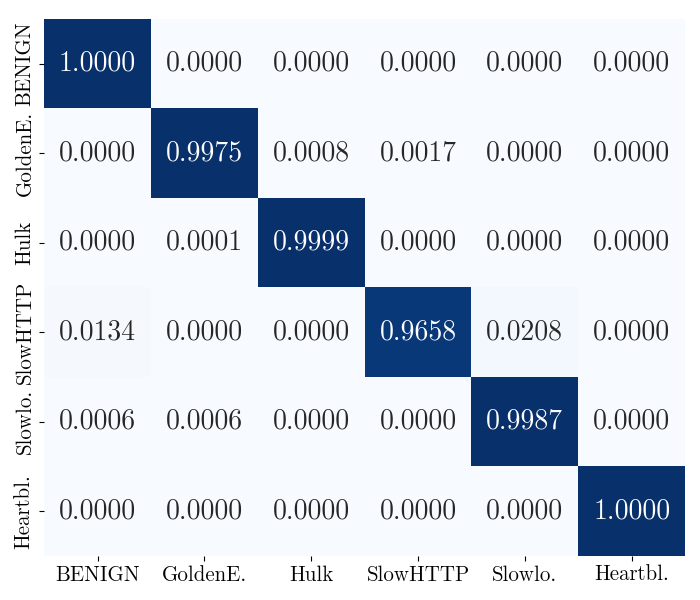} 
        & \includegraphics[width=0.2\linewidth]{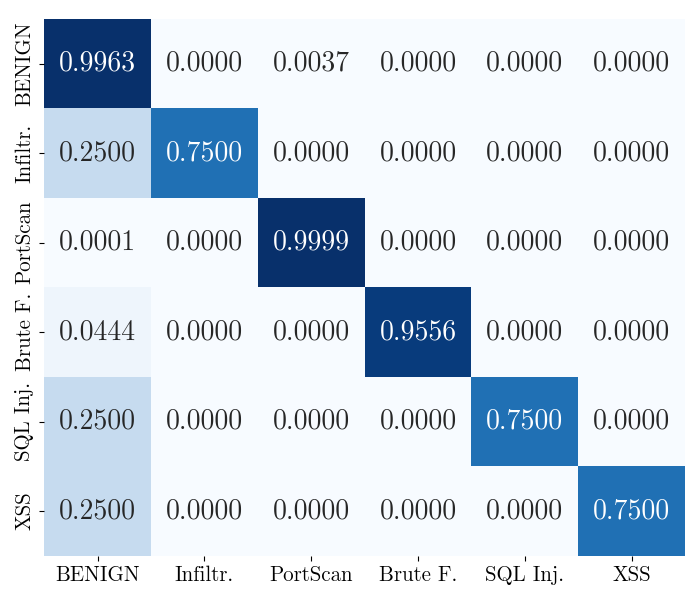} 
        & \includegraphics[width=0.2\linewidth]{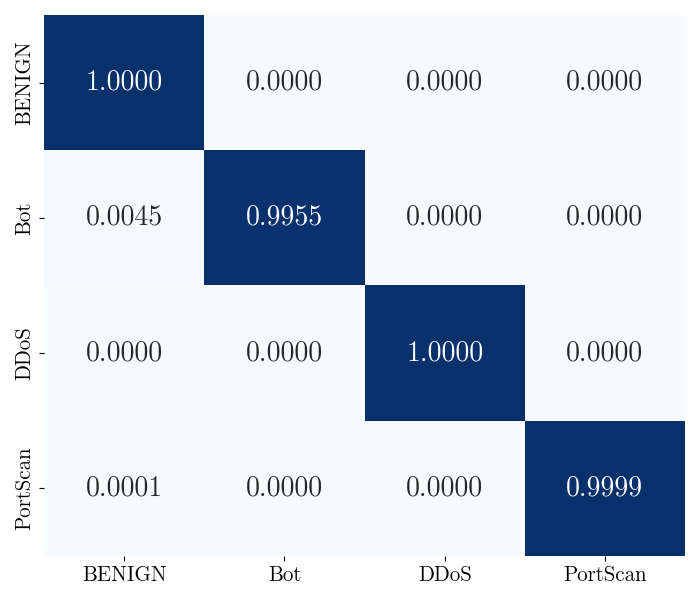}
        \\
        \rotatebox[origin=l]{\myrotation}{~~~~~NFS-2023-TE}  
        & \includegraphics[width=0.2\linewidth]{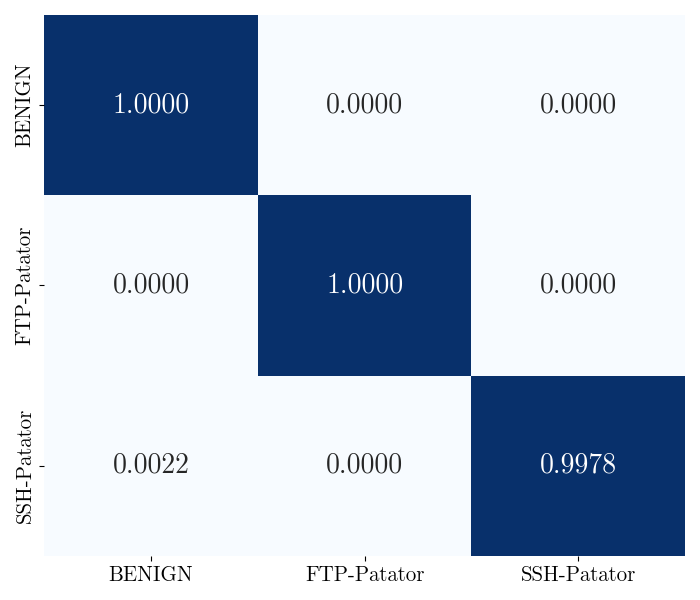} 
        & \includegraphics[width=0.2\linewidth]{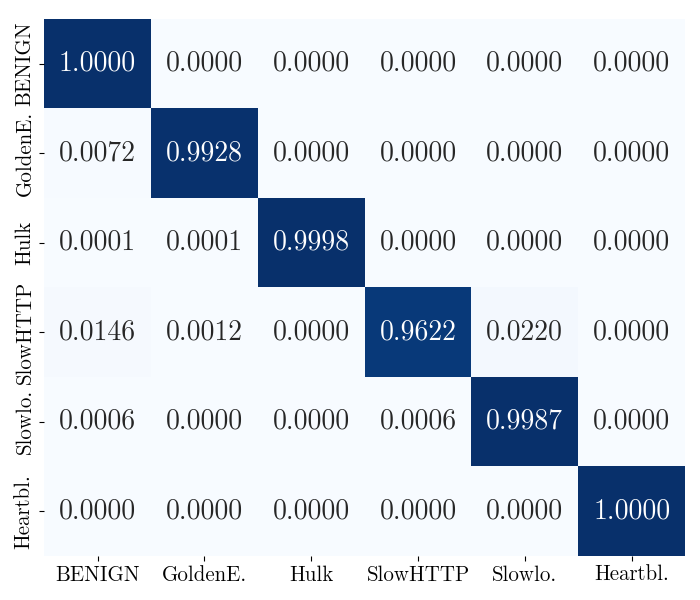} 
        & \includegraphics[width=0.2\linewidth]{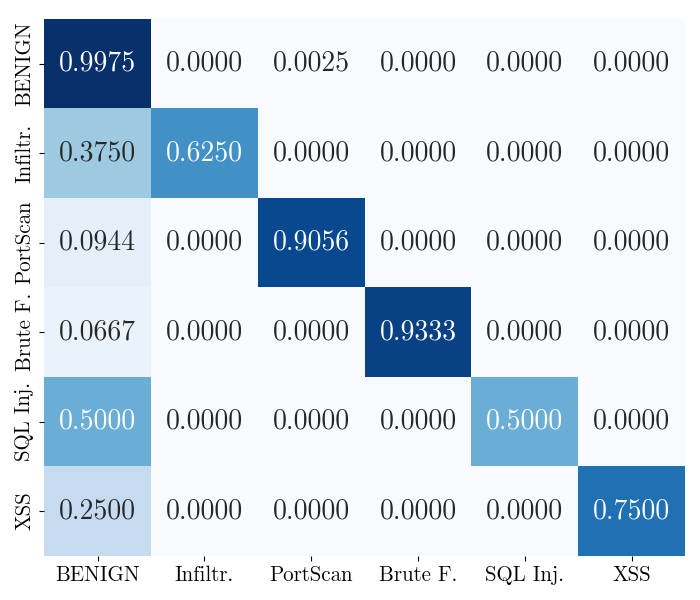} 
        & \includegraphics[width=0.2\linewidth]{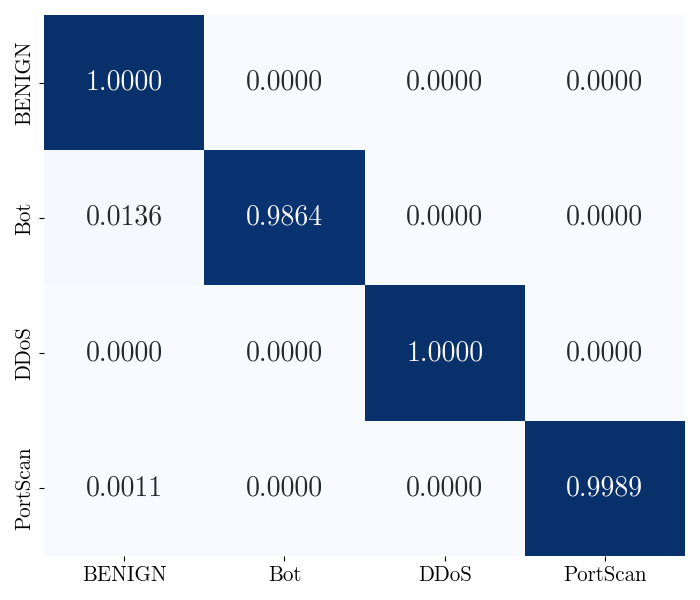}
        \\
    \end{tabular*}
\end{table*}

\section{Discussion}
\label{sec:discussion}

\subsection{Implications of Dataset Quality on RF Model Efficacy}

Upon analyzing the results from both binary and multi-class classification tables, it is evident that the RF models exhibited robust performance across various dataset versions, demonstrating high accuracy and precision in both binary and multi-class classification scenarios. This suggests that while dataset refinement and remeasurement can lead to improvements, even datasets with known biases and measurement issues can yield models with impressive performance.

However, the consistently high performance across datasets raises a critical point of discussion. It illustrates the adaptability of ML algorithms, particularly Random Forest, to learn from data with varying quality levels. Yet, it also suggests a potential risk: \textit{that high performance may mask underlying dataset flaws}. While the refinements and NFStream-generated datasets do show some improvements, the significance of these improvements is less pronounced than one might expect, given the already high performance of models trained on the original dataset. 

The RF algorithm operates by creating an ensemble of decision trees. These trees are constructed based on the statistical characteristics of the flows associated with a target label (BENIGN or a specific anomaly). Even if the statistical characteristics of the datasets vary, the RF model can still learn to classify the target label, albeit with slightly different parameters that determine the splits in the trees. From this perspective, the observation of consistent performance is not surprising, as the algorithm effectively handles its classification task. However, such results can indeed obscure underlying data quality issues.

If the flow records are not measured in a standardized manner or contain methodological errors, the resulting flow records will still exhibit certain statistical characteristics. These characteristics are sufficient for training a model, but whether the flow records accurately reflect the true nature of the network flows is a different matter. Consequently, high model performance might not necessarily indicate high data quality, but rather the model's ability to learn patterns from the available data.

Another observation is that our findings do not corroborate the substantial performance improvements reported in WTMC-2021~\cite{engelen2021} and CRiSIS-2022~\cite{lanvin2023}, particularly when these are juxtaposed with the original performance metrics of CICIDS-2017~\cite{CICIDS2017}. However, these discrepancies may stem from the intricacies involved in applying machine learning algorithms. Further study is necessary to more comprehensively understand the underlying causes of these discrepancies. 

\subsection{Multi-Model Performance Comparison}

\begin{table*}
\caption{Performance Metrics of DT, RF, and NB Algorithms Across Different Datasets and Classification Scenarios}
\label{tab:comp-models}
\fontsize{6pt}{7pt}\selectfont
  \begin{minipage}[b]{0.20\textwidth}
    \centering
\begin{tabular}{@{}lll@{}}
& & \\
& & \\
\toprule
& Day       & Model         \\ \midrule

\multirow{12}{*}{\rotatebox[origin=c]{90}{{CICIDS-2017}}} 

& Tuesday   & DecisionTree  \\
&           & RandomForest  \\
&           & NaiveBayes    \\
& Wednesday & DecisionTree  \\
&           & RandomForest  \\
&           & NaiveBayes    \\
& Thursday  & DecisionTree  \\
&           & RandomForest  \\
&           & NaiveBayes    \\
& Friday    & DecisionTree  \\
&           & RandomForest  \\
&           & NaiveBayes    \\ 
& & \\ \bottomrule
\end{tabular}
  \end{minipage}
  \hfill 
  \begin{minipage}[b]{0.38\textwidth}
    \centering
\begin{tabular}{@{}HHlllll@{}}
\multicolumn{7}{c}{Binary classification} \\
& & & & & & \\
\toprule
Day       & Model         & Precision & Recall   & Accuracy & F1 Score &  ROC AUC  \\ \midrule
Tuesday   & DecisionTree  & 0.996870  & 0.997590 & 0.999828 & 0.997230 & 0.998745 \\
          & RandomForest  & 1.000000  & 0.997108 & 0.999910 & 0.998552 & 0.998554 \\
          & NaiveBayes    & \cellcolor{lightgray}{0.045296}  & 0.998554 & \cellcolor{lightgray}{0.346642} & \cellcolor{lightgray}{0.086660} & \cellcolor{lightgray}{0.662156} \\
Wednesday & DecisionTree  & 0.998109  & 0.999404 & 0.999094 & 0.998756 & 0.999160 \\
          & RandomForest  & 0.996436  & 0.999603 & 0.998554 & 0.998017 & 0.998778 \\
          & NaiveBayes    & \cellcolor{lightgray}{0.776975}  & \cellcolor{lightgray}{0.674828} & \cellcolor{lightgray}{0.811090} & \cellcolor{lightgray}{0.722308} & \cellcolor{lightgray}{0.781965} \\
Thursday  & DecisionTree  & 0.971471  & 0.972932 & 0.999731 & 0.972201 & 0.986397 \\
          & RandomForest  & 0.996845  & 0.950376 & 0.999746 & 0.973056 & 0.975181 \\
          & NaiveBayes    & \cellcolor{lightgray}{0.019946}  & 0.917293 & \cellcolor{lightgray}{0.781761} & \cellcolor{lightgray}{0.039044} & \cellcolor{lightgray}{0.849198} \\
Friday    & DecisionTree  & 0.996368  & 0.997565 & 0.997505 & 0.996966 & 0.997514 \\
          & RandomForest  & 0.996517  & 0.997461 & 0.997524 & 0.996989 & 0.997514 \\
          & NaiveBayes    & \cellcolor{lightgray}{0.684257}  & \cellcolor{lightgray}{0.836350} & \cellcolor{lightgray}{0.774149} & \cellcolor{lightgray}{0.752697} & \cellcolor{lightgray}{0.783552} \\ 
& & & & & & \\ \bottomrule
\end{tabular}
  \end{minipage}
  \hfill 
  \begin{minipage}[b]{0.38\textwidth}
    \centering
\begin{tabular}{@{}HHlllll@{}}
\multicolumn{7}{c}{Multi-class classification} \\
& & & & & & \\
\toprule
Day       & Model         & Precision & Recall   & Accuracy & F1 Score &  ROC AUC  \\ \midrule
Tuesday   & DecisionTree  & 0.999888  & 0.999888 & 0.999888 & 0.999888 & 0.999247 \\
          & RandomForest  & 0.999925  & 0.999925 & 0.999925 & 0.999925 & 1.000000 \\
          & NaiveBayes    & 0.979395  & 0.921044 & 0.921044 & 0.943105 & 0.990472 \\
Wednesday & DecisionTree  & 0.998951  & 0.998949 & 0.998949 & 0.998950 & 0.999254 \\
          & RandomForest  & 0.999124  & 0.999123 & 0.999123 & 0.999123 & 0.999910 \\
          & NaiveBayes    & \cellcolor{lightgray}{0.858908}  & \cellcolor{lightgray}{0.566116} & \cellcolor{lightgray}{0.566116} & \cellcolor{lightgray}{0.669229} & 0.904365 \\
Thursday  & DecisionTree  & 0.998149  & 0.998081 & 0.998081 & 0.998114 & 0.990808 \\
          & RandomForest  & 0.998232  & 0.998379 & 0.998379 & 0.998269 & 0.998253 \\
          & NaiveBayes    & 0.995396  & \cellcolor{lightgray}{0.618041} & \cellcolor{lightgray}{0.618041} & \cellcolor{lightgray}{0.761719} & 0.961763 \\
Friday    & DecisionTree  & 0.997472  & 0.997548 & 0.997548 & 0.997488 & 0.999800 \\
          & RandomForest  & 0.997554  & 0.997628 & 0.997628 & 0.997568 & 0.999892 \\
          & NaiveBayes    & 0.921179  & \cellcolor{lightgray}{0.669086} & \cellcolor{lightgray}{0.669086} & \cellcolor{lightgray}{0.756887} & 0.956777 \\ 
& & & & & & \\ \bottomrule
\end{tabular}
  \end{minipage}\\
  \begin{minipage}[b]{0.20\textwidth}
    \centering
\begin{tabular}{@{}lll@{}}
& \phantom{Day}       & \phantom{Model}         \\ 
\multirow{12}{*}{\rotatebox[origin=c]{90}{{WTMC-2021}}} 

& Tuesday   & DecisionTree  \\
&           & RandomForest  \\
&           & NaiveBayes    \\
& Wednesday & DecisionTree  \\
&           & RandomForest  \\
&           & NaiveBayes    \\
& Thursday  & DecisionTree  \\
&           & RandomForest  \\
&           & NaiveBayes    \\
& Friday    & DecisionTree  \\
&           & RandomForest  \\
&           & NaiveBayes    \\ 
& & \\ \bottomrule
\end{tabular}
  \end{minipage}
  \hfill 
  \begin{minipage}[b]{0.38\textwidth}
    \centering
\begin{tabular}{@{}HHlllll@{}}
\phantom{Day}       & \phantom{Model}         & \phantom{Precision} & \phantom{Recall}   & \phantom{Accuracy} & \phantom{F1 Score} & \phantom{ROC AUC}  \\ 
Tuesday   & DecisionTree  & 0.996174  & 0.998562 & 0.999886 & 0.997367 & 0.999239 \\
          & RandomForest  & 1.000000  & 0.999521 & 0.999990 & 0.999760 & 0.999760 \\
          & NaiveBayes    & 0.946959  & 0.992809 & 0.998644 & 0.969342 & 0.995791 \\
Wednesday & DecisionTree  & 0.999534  & 0.999884 & 0.999796 & 0.999709 & 0.999816 \\
          & RandomForest  & 0.999884  & 0.999690 & 0.999851 & 0.999787 & 0.999813 \\
          & NaiveBayes    & 0.920835  & 0.964860 & 0.958681 & 0.942334 & 0.960108 \\
Thursday  & DecisionTree  & 0.948276  & \cellcolor{lightgray}{0.820896} & 0.999861 & \cellcolor{lightgray}{0.880000} & 0.910434 \\
          & RandomForest  & 1.000000  & \cellcolor{lightgray}{0.865672} & 0.999917 & 0.928000 & 0.932836 \\
          & NaiveBayes    & \cellcolor{lightgray}{0.037064}  & \cellcolor{lightgray}{0.761194} & 0.987590 & \cellcolor{lightgray}{0.070686} & \cellcolor{lightgray}{0.874462} \\
Friday    & DecisionTree  & 0.999438  & 0.999856 & 0.999670 & 0.999647 & 0.999682 \\
          & RandomForest  & 0.999490  & 0.999882 & 0.999707 & 0.999686 & 0.999718 \\
          & NaiveBayes    & 0.937509  & \cellcolor{lightgray}{0.689627} & \cellcolor{lightgray}{0.833681} & \cellcolor{lightgray}{0.794686} & \cellcolor{lightgray}{0.824696} \\ 
& & & & & & \\ \bottomrule
\end{tabular}
  \end{minipage}
  \hfill 
  \begin{minipage}[b]{0.38\textwidth}
    \centering
\begin{tabular}{@{}HHlllll@{}}
\phantom{Day}       & \phantom{Model}         & \phantom{Precision} & \phantom{Recall}   & \phantom{Accuracy} & \phantom{F1 Score} & \phantom{ROC AUC}  \\ 
Tuesday   & DecisionTree  & 0.999865  & 0.999865 & 0.999865 & 0.999865 & 0.998056 \\
          & RandomForest  & 0.999928  & 0.999928 & 0.999928 & 0.999927 & 0.999758 \\
          & NaiveBayes    & 0.999721  & 0.999720 & 0.999720 & 0.999719 & 0.994333 \\
Wednesday & DecisionTree  & 0.999676  & 0.999674 & 0.999674 & 0.999674 & 0.999729 \\
          & RandomForest  & 0.999831  & 0.999830 & 0.999830 & 0.999830 & 0.999986 \\
          & NaiveBayes    & 0.974295  & \cellcolor{lightgray}{0.777790} & \cellcolor{lightgray}{0.777790} & \cellcolor{lightgray}{0.851050} & 0.995702 \\
Thursday  & DecisionTree  & 0.999741  & 0.999741 & 0.999741 & 0.999734 & 0.902916 \\
          & RandomForest  & 0.999889  & 0.999889 & 0.999889 & 0.999881 & 0.984951 \\
          & NaiveBayes    & 0.999337  & 0.976226 & 0.976226 & 0.987490 & 0.965879 \\
Friday    & DecisionTree  & 0.999872  & 0.999872 & 0.999872 & 0.999872 & 0.999906 \\
          & RandomForest  & 0.999915  & 0.999915 & 0.999915 & 0.999915 & 0.999973 \\
          & NaiveBayes    & 0.981834  & 0.978551 & 0.978551 & 0.979659 & 0.989006 \\ 
& & & & & & \\ \bottomrule
\end{tabular}
  \end{minipage}\\
  \begin{minipage}[b]{0.20\textwidth}
    \centering
\begin{tabular}{@{}lll@{}}
& \phantom{Day}       & \phantom{Model}         \\ 
\multirow{12}{*}{\rotatebox[origin=c]{90}{{CRiSIS-2022}}} 

& Tuesday   & DecisionTree  \\
&           & RandomForest  \\
&           & NaiveBayes    \\
& Wednesday & DecisionTree  \\
&           & RandomForest  \\
&           & NaiveBayes    \\
& Thursday  & DecisionTree  \\
&           & RandomForest  \\
&           & NaiveBayes    \\
& Friday    & DecisionTree  \\
&           & RandomForest  \\
&           & NaiveBayes    \\ 
& & \\ \bottomrule
\end{tabular}
  \end{minipage}
  \hfill 
  \begin{minipage}[b]{0.38\textwidth}
    \centering
\begin{tabular}{@{}HHlllll@{}}
\phantom{Day}       & \phantom{Model}         & \phantom{Precision} & \phantom{Recall}   & \phantom{Accuracy} & \phantom{F1 Score} & \phantom{ROC AUC}  \\ 
Tuesday   & DecisionTree  & 0.996174  & 0.998562 & 0.999885 & 0.997367 & 0.999238 \\
          & RandomForest  & 1.000000  & 0.996644 & 0.999927 & 0.998319 & 0.998322 \\
          & NaiveBayes    & 0.960240  & 0.995686 & 0.999011 & 0.977642 & 0.997385 \\
Wednesday & DecisionTree  & 0.999554  & 0.999690 & 0.999737 & 0.999622 & 0.999726 \\
          & RandomForest  & 0.999922  & 0.999825 & 0.999912 & 0.999874 & 0.999892 \\
          & NaiveBayes    & 0.920418  & 0.964744 & 0.958801 & 0.942060 & 0.960193 \\
Thursday  & DecisionTree  & 0.967551  & 0.997596 & 0.996298 & 0.982344 & 0.996872 \\
          & RandomForest  & 0.968601  & 0.998097 & 0.996463 & 0.983128 & 0.997186 \\
          & NaiveBayes    & 0.950554  & 0.987879 & 0.993444 & 0.968857 & 0.990982 \\
Friday    & DecisionTree  & 0.999281  & 0.999895 & 0.999615 & 0.999588 & 0.999633 \\
          & RandomForest  & 0.999386  & 0.999908 & 0.999670 & 0.999647 & 0.999685 \\
          & NaiveBayes    & 0.957212  & 0.996574 & 0.977593 & 0.976497 & 0.978766 \\ 
& & & & & & \\ \bottomrule
\end{tabular}
  \end{minipage}
  \hfill 
  \begin{minipage}[b]{0.38\textwidth}
    \centering
\begin{tabular}{@{}HHlllll@{}}
\phantom{Day}       & \phantom{Model}         & \phantom{Precision} & \phantom{Recall}   & \phantom{Accuracy} & \phantom{F1 Score} & \phantom{ROC AUC}  \\ 
Tuesday   & DecisionTree  & 0.999927  & 0.999927 & 0.999927 & 0.999927 & 0.998791 \\
          & RandomForest  & 0.999906  & 0.999906 & 0.999906 & 0.999906 & 0.999994 \\
          & NaiveBayes    & 0.999761  & 0.999761 & 0.999761 & 0.999760 & 0.995271 \\
Wednesday & DecisionTree  & 0.999718  & 0.999717 & 0.999717 & 0.999717 & 0.999731 \\
          & RandomForest  & 0.999906  & 0.999906 & 0.999906 & 0.999906 & 0.999983 \\
          & NaiveBayes    & 0.974470  & \cellcolor{lightgray}{0.761719} & \cellcolor{lightgray}{0.761719} & \cellcolor{lightgray}{0.839700} & 0.995606 \\
Thursday  & DecisionTree  & 0.996456  & 0.996360 & 0.996360 & 0.996379 & 0.997761 \\
          & RandomForest  & 0.996535  & 0.996474 & 0.996474 & 0.996471 & 0.998615 \\
          & NaiveBayes    & 0.993587  & \cellcolor{lightgray}{0.889028} & \cellcolor{lightgray}{0.889028} & 0.937390 & 0.948438 \\
Friday    & DecisionTree  & 0.998718  & 0.998718 & 0.998718 & 0.998717 & 0.999842 \\
          & RandomForest  & 0.998736  & 0.998736 & 0.998736 & 0.998736 & 0.999908 \\
          & NaiveBayes    & 0.979859  & 0.977471 & 0.977471 & 0.978151 & 0.989379 \\ 
& & & & & & \\ \bottomrule
\end{tabular}
  \end{minipage}\\
  \begin{minipage}[b]{0.20\textwidth}
    \centering
\begin{tabular}{@{}lll@{}}
& \phantom{Day}       & \phantom{Model}         \\ 
\multirow{12}{*}{\rotatebox[origin=c]{90}{{NFS-2023-nTE}}} 
& Tuesday   & DecisionTree  \\
&           & RandomForest  \\
&           & NaiveBayes    \\
& Wednesday & DecisionTree  \\
&           & RandomForest  \\
&           & NaiveBayes    \\
& Thursday  & DecisionTree  \\
&           & RandomForest  \\
&           & NaiveBayes    \\
& Friday    & DecisionTree  \\
&           & RandomForest  \\
&           & NaiveBayes    \\ 
& & \\ \bottomrule
\end{tabular}

  \end{minipage}
  \hfill 
  \begin{minipage}[b]{0.38\textwidth}
    \centering
\begin{tabular}{@{}HHlllll@{}}
\phantom{Day}       & \phantom{Model}         & \phantom{Precision} & \phantom{Recall}   & \phantom{Accuracy} & \phantom{F1 Score} & \phantom{ROC AUC}  \\ 
Tuesday   & DecisionTree  & 0.996659  & 0.998088 & 0.999887 & 0.997373 & 0.999007 \\
          & RandomForest  & 1.000000  & 0.997132 & 0.999939 & 0.998564 & 0.998566 \\
          & NaiveBayes    & 0.954483  & 0.992352 & 0.998823 & 0.973049 & 0.995658 \\
Wednesday & DecisionTree  & 0.999444  & 0.999578 & 0.999662 & 0.999511 & 0.999642 \\
          & RandomForest  & 0.999904  & 0.999732 & 0.999874 & 0.999818 & 0.999840 \\
          & NaiveBayes    & \cellcolor{lightgray}{0.893609}  & 0.953414 & 0.944674 & 0.922543 & 0.946737 \\
Thursday  & DecisionTree  & 0.983092  & 0.999636 & 0.996856 & 0.991295 & 0.997943 \\
          & RandomForest  & 0.983391  & 0.999533 & 0.996893 & 0.991396 & 0.997925 \\
          & NaiveBayes    & 0.980334  & 0.994286 & 0.995404 & 0.987261 & 0.994967 \\
Friday    & DecisionTree  & 0.999226  & 0.999947 & 0.999616 & 0.999587 & 0.999639 \\
          & RandomForest  & 0.999344  & 0.999961 & 0.999677 & 0.999652 & 0.999696 \\
          & NaiveBayes    & 0.990988  & 0.987643 & 0.990101 & 0.989313 & 0.989935 \\ 
& & & & & & \\ \bottomrule
\end{tabular}
  \end{minipage}
  \hfill 
  \begin{minipage}[b]{0.38\textwidth}
    \centering
\begin{tabular}{@{}HHlllll@{}}
\phantom{Day}       & \phantom{Model}         & \phantom{Precision} & \phantom{Recall}   & \phantom{Accuracy} & \phantom{F1 Score} & \phantom{ROC AUC}  \\ 
Tuesday   & DecisionTree  & 0.999868  & 0.999867 & 0.999867 & 0.999867 & 0.999465 \\
          & RandomForest  & 0.999980  & 0.999980 & 0.999980 & 0.999980 & 1.000000 \\
          & NaiveBayes    & 0.999683  & 0.999683 & 0.999683 & 0.999682 & 0.993182 \\
Wednesday & DecisionTree  & 0.999484  & 0.999483 & 0.999483 & 0.999481 & 0.999674 \\
          & RandomForest  & 0.999689  & 0.999689 & 0.999689 & 0.999687 & 0.999969 \\
          & NaiveBayes    & 0.963660  & \cellcolor{lightgray}{0.785315} & \cellcolor{lightgray}{0.785315} & \cellcolor{lightgray}{0.849724} & 0.993873 \\
Thursday  & DecisionTree  & 0.996673  & 0.996623 & 0.996623 & 0.996632 & 0.998748 \\
          & RandomForest  & 0.996800  & 0.996753 & 0.996753 & 0.996755 & 0.999024 \\
          & NaiveBayes    & 0.996246  & 0.901117 & 0.901117 & 0.945372 & 0.923197 \\
Friday    & DecisionTree  & 0.999805  & 0.999805 & 0.999805 & 0.999805 & 0.999823 \\
          & RandomForest  & 0.999945  & 0.999945 & 0.999945 & 0.999945 & 0.999974 \\
          & NaiveBayes    & 0.988421  & 0.985075 & 0.985075 & 0.986387 & 0.994995 \\ 
& & & & & & \\ \bottomrule
\end{tabular}
  \end{minipage}\\
  \begin{minipage}[b]{0.20\textwidth}
    \centering
\begin{tabular}{@{}lll@{}}
& \phantom{Day}       & \phantom{Model}         \\ 
\multirow{12}{*}{\rotatebox[origin=c]{90}{{NFS-2023-TE}}} 

& Tuesday   & DecisionTree  \\
&           & RandomForest  \\
&           & NaiveBayes    \\
& Wednesday & DecisionTree  \\
&           & RandomForest  \\
&           & NaiveBayes    \\
& Thursday  & DecisionTree  \\
&           & RandomForest  \\
&           & NaiveBayes    \\
& Friday    & DecisionTree  \\
&           & RandomForest  \\
&           & NaiveBayes    \\ \bottomrule
\end{tabular}
  \end{minipage}
  \hfill 
  \begin{minipage}[b]{0.38\textwidth}
    \centering
\begin{tabular}{@{}HHlllll@{}}
\phantom{Day}       & \phantom{Model}         & \phantom{Precision} & \phantom{Recall}   & \phantom{Accuracy} & \phantom{F1 Score} & \phantom{ROC AUC}  \\ 
Tuesday   & DecisionTree  & 0.998478  & 0.999391 & 0.999950 & 0.998934 & 0.999677 \\
          & RandomForest  & 0.999087  & 0.999695 & 0.999971 & 0.999391 & 0.999837 \\
          & NaiveBayes    & \cellcolor{lightgray}{0.055769}  & 0.999086 & \cellcolor{lightgray}{0.603808} & \cellcolor{lightgray}{0.105641} & \cellcolor{lightgray}{0.796707} \\
Wednesday & DecisionTree  & 0.999369  & 0.999178 & 0.999707 & 0.999274 & 0.999509 \\
          & RandomForest  & 0.999522  & 0.999293 & 0.999761 & 0.999407 & 0.999586 \\
          & NaiveBayes    & \cellcolor{lightgray}{0.864749}  & 0.952102 & 0.960339 & 0.906325 & 0.957260 \\
Thursday  & DecisionTree  & 0.983588  & \cellcolor{lightgray}{0.899925} & 0.983046 & 0.939899 & 0.948666 \\
          & RandomForest  & 0.984192  & 0.900019 & 0.983142 & 0.940225 & 0.948761 \\
          & NaiveBayes    & \cellcolor{lightgray}{0.477772}  & 0.987316 & \cellcolor{lightgray}{0.839154} & \cellcolor{lightgray}{0.643937} & 0.900437 \\
Friday    & DecisionTree  & 0.999247  & 0.999247 & 0.999502 & 0.999247 & 0.999437 \\
          & RandomForest  & 0.999312  & 0.999247 & 0.999523 & 0.999279 & 0.999453 \\
          & NaiveBayes    & \cellcolor{lightgray}{0.854009}  & 0.987666 & 0.940037 & 0.915987 & 0.952070 \\ \bottomrule
\end{tabular}
  \end{minipage}
  \hfill 
  \begin{minipage}[b]{0.38\textwidth}
    \centering
\begin{tabular}{@{}HHlllll@{}}
\phantom{Day}       & \phantom{Model}         & \phantom{Precision} & \phantom{Recall}   & \phantom{Accuracy} & \phantom{F1 Score} & \phantom{ROC AUC}  \\ 
Tuesday   & DecisionTree  & 0.999957  & 0.999957 & 0.999957 & 0.999957 & 0.999838 \\
          & RandomForest  & 0.999971  & 0.999971 & 0.999971 & 0.999971 & 0.999998 \\
          & NaiveBayes    & 0.985369  & \cellcolor{lightgray}{0.898914} & \cellcolor{lightgray}{0.898914} & 0.933937 & 0.998703 \\
Wednesday & DecisionTree  & 0.999484  & 0.999484 & 0.999484 & 0.999482 & 0.999609 \\
          & RandomForest  & 0.999596  & 0.999596 & 0.999596 & 0.999594 & 0.999938 \\
          & NaiveBayes    & 0.973447  & \cellcolor{lightgray}{0.502993} & \cellcolor{lightgray}{0.502993} & \cellcolor{lightgray}{0.626339} & 0.994136 \\
Thursday  & DecisionTree  & 0.983512  & 0.983479 & 0.983479 & 0.983190 & 0.995042 \\
          & RandomForest  & 0.983567  & 0.983554 & 0.983554 & 0.983249 & 0.995329 \\
          & NaiveBayes    & 0.924668  & \cellcolor{lightgray}{0.676976} & \cellcolor{lightgray}{0.676976} & \cellcolor{lightgray}{0.750929} & 0.914001 \\
Friday    & DecisionTree  & 0.999489  & 0.999489 & 0.999489 & 0.999489 & 0.999849 \\
          & RandomForest  & 0.999519  & 0.999519 & 0.999519 & 0.999519 & 0.999926 \\
          & NaiveBayes    & 0.973774  & 0.968256 & 0.968256 & 0.969951 & 0.987969 \\ \bottomrule
\end{tabular}
  \end{minipage}
\end{table*}

We also evaluated the performance of the Decision Tree (DT) and Naive Bayes (NB) algorithms, benchmarking them against the RF model. For model training, the top 15 features were selected using the Extremely Randomized Trees (Extra-Trees) algorithm~\cite{Geurts_2006}. \Cref{tab:comp-models} compares the achieved performance across various datasets and classification scenarios. Metrics below 0.9 have been highlighted in gray. From \Cref{tab:comp-models}, several key conclusions can be derived.

Across both binary and multi-class classifications, RF consistently outperforms DT and NB. This superior performance is reflected in its near-perfect precision, recall, accuracy, F1 score, and AUC ROC. The robustness of RF is attributed to its ensemble nature, which aggregates multiple decision trees, thereby reducing overfitting and enhancing generalization.

DT demonstrates strong performance in most scenarios but exhibits more variability compared to RF. Notably, its lower performance on the Thursday datasets in binary classification highlights its sensitivity to data variations and potential overfitting to specific features.

NB shows the most significant fluctuations in performance, particularly in precision and recall. This variability stems from NB's assumption of feature independence, which may not hold true in complex datasets, resulting in higher false positive rates and lower overall accuracy.

The day-to-day performance variability, especially in NB, underscores the impact of data distribution on model performance. The consistency of RF suggests it is better equipped to handle varying data distributions, while DT and NB are more sensitive to such changes.

From \Cref{tab:comp-models}, we also find that the best performance was achieved on the CRiSIS-2022 and NFS-2023-nTE datasets. Here, the accuracies are consistently high across both binary and multi-class classification. This suggests that the CRiSIS-2022 dataset and our dataset with no TCP flow expiration policy applied are particularly well-suited for these models.

Across the other three datasets, CICIDS-2017, WTMC-2021, and NFS-2023-TE, several metrics perform very poorly, especially for the NB algorithm. Notably, there are cases where precision, accuracy, F1 score, and ROC AUC are poor, but recall is close to 100\%. This suggests that while the model identifies most positive cases, it does so with a high rate of false positives, particularly for NB. This highlights the need for careful feature selection and model tuning to improve performance in these datasets.

In conclusion, the high performance of RF in both binary and multi-class scenarios makes it a preferred choice for anomaly detection tasks, where robust and consistent performance is crucial. DT can serve as a good alternative when interpretability and lower computational cost are prioritized, despite its occasional variability. NB, while less reliable overall, can be useful in scenarios with well-understood and consistent data distributions.

\subsection{Conclusive Insights}

The examination of the results leads us to a nuanced understanding. On one hand, the resilience of ML models, particularly RF, to dataset imperfections is encouraging, as it demonstrates their ability to effectively capture underlying patterns of anomalous behavior. RF's consistent performance across datasets, despite variations in data quality, highlights its robustness and adaptability.

However, this robustness also underscores the critical need for vigilance in dataset creation. The consistently high performance of RF models can mask underlying data quality issues, as high model performance might not necessarily indicate high data quality, but rather the model's ability to learn patterns from the available data. \textit{This emphasizes the importance of refined measurement techniques. These techniques are essential not because they drastically change model performance metrics, but because they ensure that the models are learning from data that accurately represents real-world conditions.} Ensuring data quality is particularly crucial when deploying these models in operational settings, where the stakes of misclassification are high.

Our comparative analysis of RF, DT, and NB models revealed several important insights. While RF consistently outperforms the other models, the performance variability in DT and especially in NB highlights the sensitivity of these models to data quality and distribution. DT exhibits strong performance but with more variability, particularly in datasets with specific attack types. NB, on the other hand, shows significant fluctuations in performance due to its assumption of feature independence, which may not hold true in complex datasets. This sensitivity to data variations further underscores the need for careful feature selection and model tuning to improve performance, especially in datasets with known biases or measurement issues.

\textit{The potential masking effect observed in RF is not as pronounced in DT and NB, which tend to be more directly affected by data quality. This difference indicates that while RF can maintain high performance even with imperfect data, DT and NB provide more visible indicators of data quality issues through their performance variability.}

In summary, our comprehensive reflection underscores the complexity of evaluating dataset quality solely based on model performance metrics. It highlights the importance of continuously striving for dataset accuracy and sound methodology to ensure that high model performance corresponds with true predictive power in real-world scenarios. Ensuring data quality and methodological rigor is essential for the effective deployment of these models in operational settings, where the consequences of misclassification can be significant.

\section{Limitations and Future Prospects}
\label{sec:limit}

\subsection{Real-time Applicability of the Datasets}

A fundamental limitation inherent in all versions of the CICIDS-2017 dataset is their constrained utility in real-world, especially real-time anomaly detection scenarios. The core issue with these datasets stems from the manner in which the flow records were generated. Predominantly, the tools were configured to produce comprehensive flow records encapsulating the complete lifecycle of network flows. While this methodology ensures detailed record-keeping, it falls short of mirroring the real-time flow characteristics that are pivotal in dynamic network environments. Essential aspects like flow size, packet count, or duration thresholds, which are critical in simulating real-world traffic patterns, are not adequately represented.

This limitation manifests in the datasets' inability to offer a realistic portrayal of the partial and evolving nature of network flow information, which is crucial for real-time monitoring and anomaly detection. In practical scenarios, anomaly detection systems often rely on incomplete or evolving flow data to function effectively and adaptively. The complete flow records in the datasets may not, therefore, offer an accurate reflection of how ML models perform in such time-sensitive settings. This discrepancy poses significant challenges to the transferability of research findings based on these datasets to practical, real-time detection systems. The research community must be cautious in extrapolating the results derived from these datasets to real-world applications, as the underlying data may not accurately simulate the conditions encountered in real-time network environments.

While NFStream offers a capability to meter flows up to specific thresholds, this aspect was not explored within the scope of the current study. Future research, therefore, should focus on utilizing NFStream or similar tools to create datasets that better reflect the partial flow information characteristic of real-time network environments. Such datasets would be invaluable in evaluating the effectiveness of ML models under conditions that demand swift and accurate anomaly detection using incomplete data. This direction of research promises to bridge the gap in understanding the applicability and performance of ML models in real-time anomaly detection scenarios, a critical step towards enhancing the practical relevance of anomaly detection methodologies.

\subsection{Scalability Considerations}

The scalability implications of this study highlight several potential limitations. While RF models have shown impressive accuracy and robustness, their computational intensity poses challenges for scalability. Deploying RF models in real-time, high-throughput environments may require significant infrastructure, including parallel processing capabilities and high-memory systems. This can limit the practical deployment of RF in resource-constrained settings.

DT and NB models, which were also analyzed in this study, exhibit their own scalability challenges. Although DT models are generally less computationally demanding than RF, they still require careful tuning and feature selection to maintain performance at scale. DT models can become inefficient when dealing with very large datasets or high-dimensional feature spaces, potentially leading to slower processing times and increased memory usage.

NB models, while computationally efficient and quick to train, showed significant fluctuations in performance. The assumption of feature independence in NB may not hold true in complex datasets, leading to scalability issues as the model may struggle to handle the nuances of large-scale data accurately. Additionally, NB’s sensitivity to data distribution changes can result in varying performance when scaled up to larger datasets.

In light of the above, future work should focus on quantifying the computational resources required for training and deploying these models in real-world settings, considering the studied classification contexts. This involves not only measuring the time and memory needed but also assessing the infrastructure costs and feasibility of deploying these models in operational environments.

\subsection{Application of Sophisticated Machine Learning}

Our research has been concentrated on the analysis of single flows, without delving into the interdependencies that might exist between different flow samples. This specific focus has made supervised learning techniques, RF, DT, and NB, more suitable for our study, in comparison to more complex methods like unsupervised learning or advanced models such as Isolation Forest (IF) and Long Short-Term Memory (LSTM) networks.

In our approach, we utilized features that represent the externally observable characteristics of network flows, predominantly derived from packet aggregation. A critical observation here is the possibility that certain attack signatures might not show substantial deviations from typical flow statistics when analyzed individually. These signatures could instead be more indicative of the nature of specific attacks, aspects that are not easily identifiable from external flow characteristics alone. As a result, while IF is known for its proficiency in outlier detection, its effectiveness in distinguishing between certain types of anomalies and normal flows might be limited. This limitation is particularly pronounced when the anomalies in question do not manifest as distinct outliers in the statistical analysis of individual flows.

Furthermore, the application of LSTM networks, which are inherently designed for time-series data, encounters unique challenges in the context of network flow records. The fundamental nature of network flows, with their varied start and end times, durations, and intervals, does not align perfectly with typical time-series characteristics. This discrepancy, further exacerbated by sporadic traffic bursts, complicates the task of organizing these flows into coherent time windows or extracting features representative of such patterns. The complexity involved in processing and analyzing network flows in a manner suitable for LSTM networks demands a different methodological approach that was beyond the scope of our current examination.


Given these considerations, a further area for future research lies in exploring and developing algorithms that can effectively process and analyze network flows, taking into account their unique characteristics and interdependencies. Such advancements are required to enhance the applicability and effectiveness of machine learning techniques in practical, dynamic network environments. 

\subsection{Implications of FIN/RST Flags in Flows}

As discussed throughout this study, we observe a considerable number of flows with FIN/RST flags (see \Cref{tbl:fin-rst}). The body of the study focused on the handling of such flags in the context of standard measurement methodologies, i.e., how a flow should be treated upon observing a packet with such a flag (see \Cref{sec:objectives}).

However, the presence of FIN/RST flags can also be indicative of the server being saturated due to the effectiveness of the attack being performed towards it. If the server is close to its saturation due to the massive number of connections, it will start terminating the connections to avoid crashing or service outages. This is manifested in the network traffic flows as packets with an increased number of FIN/RST flags.

This observation raises an interesting dilemma: whether to include these packets in the flows or exclude them via TCP flag-based expiration policies (see \Cref{sec:TCP-policy}), as applied during the generation of the NFS-2023-TE dataset. Technically, these packets are the product of the attack and might not be mixed with regular attack signatures.

Including these FIN/RST packets in the flow records provides a more comprehensive view of the network's response to the attack, reflecting both the attack traffic and the server's defensive measures. This approach ensures that the dataset captures the full spectrum of network behavior under stress, which can be invaluable for understanding the dynamics of the attack and the server's response. Moreover, this comprehensive approach can improve the robustness of ML models by training them on a dataset that includes various aspects of attack scenarios, including the server's attempts to mitigate the impact.

On the other hand, excluding these packets through TCP flag-based expiration policies could help isolate the attack signatures more clearly, avoiding potential noise introduced by the server's defensive actions. This can lead to cleaner datasets that focus strictly on the attack patterns, which might be preferable for certain types of analysis or for training models specifically aimed at detecting the initial stages of an attack before server saturation occurs.

In our study, NFS-2023-nTE does not implement TCP flag expiry, while NFS-2023-TE does. However, both the single model evaluation and multi-model evaluation did not show noticeable performance differences across these two datasets. This suggests that for the models assessed, taking into account TCP FIN/RST flags did not significantly impact the classification performance. Nonetheless, further research is needed to extend the examination to a wider range of machine learning algorithms. Future research could explore hybrid approaches that leverage the strengths of both methodologies, potentially offering a more nuanced understanding of network behaviors in the presence of attacks.



\subsection{Handling of Flows with Zero Packet Payloads}

This study marked flows with zero packet payload as BENIGN despite these flows being originally labeled as attacks by the dataset authors~\cite{icissp17}. This decision is based on the observation that the attack types used in the traffic traces typically involve data transfer. Flows with no data being transmitted, although originally labeled as attacks, do not exhibit the characteristics of real-world attacks since the flow features derived from packet payload sizes (e.g., minimum, maximum, mean, and standard deviation of packet sizes in forward and backward directions, or flow size) remain zero. This can potentially introduce bias into model development. Therefore, we relabeled such flows as BENIGN traffic. This approach is consistent with the strategy applied in~\cite{engelen2021}, where the authors also relabeled such flows due to similar considerations.

However, our evaluations demonstrated that this relabeling strategy did not significantly impact model performance. Notably, the CICIDS-2017 dataset, which did not implement this relabeling strategy, showed performance metrics comparable to those of the other datasets. This finding suggest that the evaluated models are resilient to the presence of flows with zero packet payloads. This resilience might be due to the models' ability to generalize from the available data and focus on features that more strongly indicate attack patterns.

To gain a better understanding of the impact of handling potentially anomalous flows with zero packet payloads, further studies are required. Future research should explore whether this strategy of relabeling impacts the detection capabilities and overall performance of models in various scenarios.


\section{Conclusion}
\label{sec:conclusion}

Our comprehensive study embarked on the quest to enhance the integrity of network traffic datasets, a cornerstone for developing machine learning models in cybersecurity anomaly detection. By deploying NFStream, we carefully crafted two refined versions of the CICIDS-2017 dataset that adhere to methodologically sound flow measurement principles. The subsequent evaluation using the RF algorithm provided a multi-faceted view of the models' performance across the original, refined, and NFStream-generated datasets.

The research unearthed several key insights. Despite data measurement inconsistencies in existing datasets, the RF model demonstrated remarkable resilience, showcasing consistent accuracy in anomaly classification. However, this robustness does not overshadow the importance of rigorous dataset preparation, as evidenced by negative values and NaN entries in datasets not processed by NFStream.

This work underscores the critical need for transparent and accurate flow metering in network traffic analysis. Future research should continue to explore the development of sophisticated flow metering tools that can seamlessly integrate into the ML pipeline, ensuring that models are trained on data reflective of the multifaceted nature of network behavior.


In alignment with the principles of open science, the refined datasets, NFS-2023-nTE~\cite{nfs-a} and NFS-2023-TE~\cite{nfs-b}, have been made accessible to the broader research community. Furthermore, we have ensured that the scripts utilized in our experiments are also publicly available~\cite{repo}. Our reproducible methodology is readily adaptable for processing the CICIDS-2017 PCAP files to meet the diverse needs of future research endeavors. This initiative aims to promote a deeper understanding of our methodological approaches and foster further research within this area of study.  





\section*{Acknowledgement}

This work was supported by the János Bolyai Research Scholarship of the Hungarian Academy of Sciences. Supported by the ÚNKP-23-5-BME-461 New National Excellence Program of the Ministry for Culture and Innovation from the source of the National Research, Development and Innovation Fund. 
The work presented in this paper was supported by project no. TKP2021-NVA-02. Project no. TKP2021-NVA-02 has been implemented with the support provided by the Ministry of Culture and Innovation of Hungary from the National Research, Development and Innovation Fund, financed under the TKP2021-NVA funding scheme.











\printcredits


\bibliography{references}

\newpage

\bio{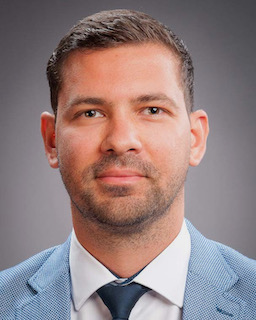}
Adrian Pekar earned his PhD from the Technical University of Kosice, Slovakia, in 2014. He presently holds the position of Associate Professor at the Department of Networked Systems and Services within the Budapest University of Technology and Economics, Hungary. Before joining academia in Hungary, he gained valuable experience through research, teaching, and engineering roles across Slovakia and in New Zealand. His research interests encompass a wide array of topics, including network and service management, software-defined networking, network function virtualization, network programmability, machine learning, and data science.
\endbio

\bio{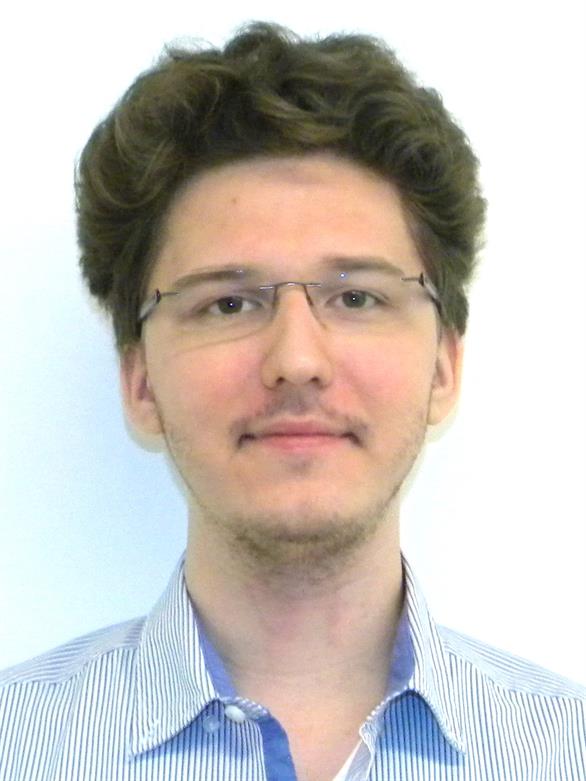}
Richárd Józsa earned his BSc degree from the Department of Networked Systems and Services at the Budapest University of Technology and Economics, Hungary, in 2023. He is currently furthering his education as an MSc student in the same department, while also serving as a Research Fellow. His research interests encompass network and service management, software-defined networking, and network function virtualization. 
\endbio

\end{document}